\documentclass[lettersize,journal]{IEEEtran}
\usepackage{amsmath,amsfonts}
\usepackage{array}
\usepackage{textcomp}
\usepackage{stfloats}
\usepackage{url}
\usepackage{verbatim}
\usepackage{graphicx}
\usepackage{cite}
\usepackage{marvosym}
\usepackage{balance}
\usepackage{subfigure}
\usepackage[vlined,ruled,linesnumbered]{algorithm2e}
\usepackage{threeparttable}
\usepackage{diagbox}
\usepackage{booktabs}
\usepackage{multirow} 
\usepackage{xcolor}
\usepackage{bm}
\newtheorem{observation}{\textbf{Observation}}
\newtheorem{proposition}{\textbf{Proposition}}
\begin{document}

\title{A Scene-aware Models Adaptation Scheme for Cross-scene Online Inference on Mobile Devices}

\author{Yunzhe~Li,~\IEEEmembership{Student Member,~IEEE}, 
	Hongzi~Zhu,~\IEEEmembership{Senior~Member,~IEEE},
	Zhuohong~Deng, Yunlong~Cheng,~\IEEEmembership{Student Member,~IEEE}, 
    Zimu~Zheng,~\IEEEmembership{Member,~IEEE},
    Liang~Zhang,~\IEEEmembership{Student Member,~IEEE}, Shan~Chang,~\IEEEmembership{Member,~IEEE}, Minyi~Guo,~\IEEEmembership{Fellow,~IEEE} \thanks{Yunzhe Li, Hongzi Zhu (corresponding author), Zhuohong Deng, Yunlong Cheng, Liang Zhang and Minyi Guo are with the Department of Computer Science and Engineering, Shanghai Jiao Tong University, Shanghai, 200240, China, E-mail: hongzi@sjtu.edu.cn;

    Zimu Zheng is with Huawei Cloud, Shenzhen, 518116, China;
 
	Shan Chang is with the School of Computer Science and Technology, Donghua University, Shanghai, 201620, China.}}
	

\markboth{Journal of \LaTeX\ Class Files,~Vol.~X, No.~X, XX~XXXX}%
{Shell \MakeLowercase{\textit{et al.}}: A Sample Article Using IEEEtran.cls for IEEE Journals}


\maketitle

\begin{abstract}
Emerging Artificial Intelligence of Things (AIoT) applications desire online prediction using deep neural network (DNN) models on mobile devices.
However, due to the movement of devices, \emph{unfamiliar} test samples constantly appear, significantly affecting the prediction accuracy of a pre-trained DNN. In addition, unstable network connection calls for local model inference.
In this paper, we propose a light-weight scheme, called \emph{Anole}, to cope with the local DNN model inference on mobile devices. 
The core idea of Anole is to first establish an army of compact DNN models, and then adaptively select the model fitting the current test sample best for online inference.
The key is to automatically identify \emph{model-friendly} scenes for training scene-specific DNN models. To this end, we design a weakly-supervised scene representation learning algorithm by combining both human heuristics and feature similarity in separating scenes. Moreover, we further train a model classifier to predict the best-fit scene-specific DNN model for each test sample. 
We implement Anole on different types of mobile devices and conduct extensive trace-driven and real-world experiments based on unmanned aerial vehicles (UAVs). The results demonstrate that Anole outwits the method of using a versatile large DNN in terms of prediction accuracy (4.5\% higher), response time (33.1\% faster) and power consumption (45.1\% lower). 
\end{abstract}


\section{Introduction}
\textbf{Motivation.} Last decade has witnessed the booming development of Artificial Intelligence of Things (AIoT), an emerging computing paradigm that marries artificial intelligence (AI) and Internet of Things (IoT) technologies to enable independent decision-making at each component level of the interconnected system.
In many AIoT scenarios, deep neural network (DNN) model inference (\emph{i.e.}, prediction) tasks are required to execute on mobile devices, referred to as the \emph{online mobile inference} (OMI) problem, with stringent accuracy and latency requirements. For example, unmanned aerial vehicles (UAVs) need to constantly detect surrounding objects in real time \cite{wang2023generalized}; a dash cam mounted on a vehicle needs to perform continuous image object detection \cite{zhou2021monoef}; robots in smart factories need to detect objects in production lines in real time, interact with human workers and other robots \cite{zheng2020contextual} or even make production decisions \cite{dalmia2020pest}.


To address the OMI problem, however, is demanding for two reasons as follows. First, given that mobile devices constantly experience scene changes while moving (\emph{e.g.}, due to various lighting conditions, weather conditions, and viewing angles), the output of DNNs should remain reliable and accurate.
Training a statistical learning DNN on a given dataset, as in normal deep learning paradigm, becomes difficult to guarantee the robustness, interpretability and correctness of the output of the statistical learning models when data samples are \emph{out-of-distribution} (OOD) \cite{chen2019edge}. 
Second, the response time for model inference should satisfy a rigid delay budget to support real-time interactions with these devices. As mobile devices are resource-constrained in terms of computation, storage and energy, they cannot handle large DNNs. Though it would be beneficial to offload a part of or even entire inference tasks to a remote cloud, unstable communication between mobile devices and the cloud may lead to unpredictable delay.

%

In the literature, much effort has been made to improve DNN model inference accuracy on mobile devices but in static scenarios. One main branch aims to develop DNNs specially designed for mobile devices~\cite{howard2017mobilenets, zhang2018shufflenet, sanh2019distilbert, jiao2019tinybert} or to compress (\emph{e.g.}, via model pruning and quantization) existing DNNs to match the computing capability of a mobile device~\cite{han2015learning, wang2020pruning}. Such schemes ensure real-time model inference at the expense of compromised accuracy, especially when dealing with OOD data samples. Moreover, even without considering the limited resources of mobile devices, it is also difficult to find a satisfactory neural network capable of stable inference on OOD samples for some complex tasks such as mobile decision making \cite{zheng2022leon, zheng2020contextual}.
Another branch is to divide DNNs and perform collaborative inference on both edge devices and the cloud~\cite{hu2019dynamic, kang2017neurosurgeon, fang2019teamnet}, or to transmit compressed sensory data to the cloud for data recovery and model inference~\cite{Liu2019, zhang2021elf}. These approaches need coordination with the cloud for each inference, leading to unpredictable inference delays when the communication link is unstable or disconnected. As a result, to the best of our knowledge, there is no successful solution to the OMI problem yet.

\textbf{Our approach.} We propose \emph{Anole}, which enables online model inference on mobile devices in dynamic scenes. We have the insight that a compressed DNN targeted for a particular scene (\emph{i.e.}, data distribution) can achieve comparable inference accuracy provided by a fully-fledged large DNN. 
The core idea of Anole is to first establish a colony of compressed scene-specific DNNs, and then adaptively select the model best suiting the current test sample for online inference.
To this end, it is essential to identify scenes from the perspective of DNN models. We design a weakly-supervised scene representation learning scheme by combining both human heuristics and feature similarity in separating scenes. After that, for each identified scene, an individual compressed DNN model can be trained. Furthermore, we train a model classifier to predict the best-fit compressed DNN models for use during online inference. 
As a result, compelling prediction accuracy can be achieved on mobile devices by actively recruiting most capable compressed models, without any intervention with the cloud.

\textbf{Challenges and contributions.} The Anole design faces three main challenges. 
First, how to obtain model-friendly scenes and train scene-specific DNNs from public datasets is unclear, as the distribution that a DNN model can characterize is implicit. 
One naive way is to use semantic attributes (\emph{e.g.}, time, location, weather and light conditions) of data to define scenes of similar data samples. However, as shown in our empirical study, DNNs trained on such scenes cannot reach satisfactory prediction accuracy even on their respective training scenes.
To tackle this challenge, we design a scene representation learning algorithm that combines semantic similarity and feature similarity of data to filter out scenes. Specifically, human heuristic is first used to define scenes of similar semantic attribute values, referred to as \emph{semantic scenes}. Then, a scene representation model, denoted as $\mathcal{M}_{scene}$, is trained using the indices of semantic scenes as labels. After that, we can obtain embeddings of all data samples extracted with $\mathcal{M}_{scene}$ and believe such embeddings can well characterize semantic information. Therefore, by conducting multi-granularity clustering on these embeddings, we can obtain clusters of data samples with similar semantic information in feature space, referred to as \emph{model-friendly scenes}. Finally, a compressed DNN can be trained on each model-friendly scene, constituting a model repository for use.

Second, given a test sample, how to determine the best-fit models or whether such models even exist in the model repository is hard to tell. To deal with this challenge, we train a model classifier, denoted as $\mathcal{M}_{decision}$, to predict the best model for use. Specifically, for each model-friendly scene, we select those data samples in the scene that can be accurately predicted by the corresponding DNN and use the index of the DNN as the label to train $\mathcal{M}_{decision}$. Instead of testing all data samples, we use Thompson sampling to establish balanced training sets at a low computational cost. With a well-trained $\mathcal{M}_{decision}$, the most suitable compressed models can be predicted and the prediction confidence can be used to indicate whether such models exist.

Last, how to deploy those pre-trained compressed DNNs on mobile devices with constrained memory is non-trivial. We have the observation that the utility of models follows a power-law distribution over all test videos. This implies that it is feasible to cache a small number of most frequently used compressed models and take a least frequently used (LFU) model replacement strategy.

We implement Anole on three typical mobile devices, \emph{i.e.}, Jetson Nano, Jetson TX2 NX and a laptop, with each equipped with a CPU/MCU and an entry-level GPU, to conduct the image object detection task on moving vehicles. Specifically, we train the $\mathcal{M}_{scene}$ based on Resnet18~\cite{he2016deep}, a pack of 19 compressed DNNs based on YOLOv3-tiny~\cite{adarsh2020yolo}, and the $\mathcal{M}_{decision}$ based on Resnet18 accordingly, using three driving video datasets collected from multiple cities in different counties. 
We conduct extensive trace-driven and real-world experiments using UAVs. Results demonstrate that Anole is lightweight and agile to switch best models with low latencies of 61.0 ms, 13.9 ms, and 52.0 ms on Jetson Nano, Jetson TX2 NX, and the laptop, respectively. In cross-scene (\emph{i.e.}, seen but fast-changing scenes) setting, Anole can achieve a high F1 prediction accuracy of 56.4\% whereas the F1 score of a general large DNN model and a general compact DNN are 50.7\% and 45.9\%, respectively. In hard new-scene (\emph{i.e.}, unseen scenes) setting, Anole can maintain a high F1 score of 48.7\% whereas the F1 score of the general large DNN and the general compact DNN drops to 46.6\% and 41.1\%, respectively.


We highlight the main contributions made in this paper as follows:

\begin{itemize}[]
	\item A new solution to the OMI problem by recruiting a pack of compact but specialized models on resource-constrained mobile devices, without any intervention with the cloud during online model inference;
	\item A scene partition method that effectively facilitates the training of specialized models by leveraging both semantic and feature similarity of the data;
	\item We have implemented Anole on typical mobile devices and conducted extensive trace-driven and real-world experiments on 3 typical tasks, the results of which demonstrate the efficacy of Anole.
\end{itemize}


\section{Problem Definition}

\subsection{System Model}
%

We consider three types of entities in the system:

\begin{itemize}
	\item \textbf{Mobile devices:} Mobile devices have constrained computational power and a limited amount of memory but are affordable for running and storing compressed DNNs. Such devices may be moving while performing online inference tasks at the same time. They are battery-powered, desiring lightweight operations. In addition, they can communicate with a cloud server via an unstable wireless network connection for offline model training and downloading.
	
	\item \textbf{Cloud server:} A cloud server has sufficient computational power and storage for offline model training. During online inference, the cloud server is not involved.
	
	\item \textbf{Complex environment}: We consider practical environments where background objects and light conditions have distinct spatial and temporal distributions. When mobile devices move in such a complex environment, they constantly experience fast scene changes.  
\end{itemize}

\subsection{Problem Formulation}
\label{subsec:PF}

Given the set of all available labeled data, denoted as $D$, a compressed DNN model, denoted as $\mathcal{M}_i$, can be trained on a particular dataset, denoted as $\Gamma_i$, which is a subset of $D$, \emph{i.e.}, $\Gamma_i \subseteq D$ for $i\in\mathbb{N}$. For instance, $\Gamma_i$ can be established based on some semantic attributes of data.
Assume that a set of $n$ models $\{\mathcal{M}_1, \mathcal{M}_2, \cdots, \mathcal{M}_n\}$ have been pre-trained on respective training datasets $\{\Gamma_1, \Gamma_2, \cdots, \Gamma_n\}$ and the \emph{implicit} data distributions that those models can characterize are $\{\Psi_1, \Psi_2, \cdots, \Psi_n\}$, respectively, which means that if a data sample $x \in \Psi_i$ for $i\in[1,n]$, model $\mathcal{M}_i$ guarantees to output accurate prediction for $x$. We have the following proposition:

\begin{proposition}
	\emph{Though $\mathcal{M}_i$ is trained on $\Gamma_i$, not all data samples in $\Gamma_i$ necessarily belong to $\Psi_i$, i.e., $\Gamma_i \not\subset \Psi_i$.}
	\label{proposition1}
\end{proposition}

\begin{figure}[]
	\centering
	\includegraphics[height=3.6cm]{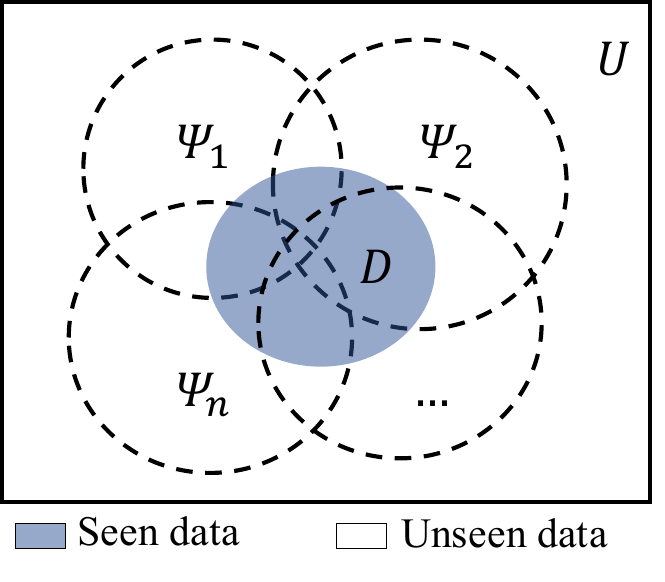}
	\caption{Illustration of the online mobile inference problem, where data distributions characterized by statistical models (depicted as dashed disks) are implicit and not easy to understand.}
	\label{fig:omi}
\end{figure}

As illustrated in Figure \ref{fig:omi}, given $D$, we can train such a set of $n$ models $M=\{\mathcal{M}_1, \mathcal{M}_2, \cdots, \mathcal{M}_n\}$ so that $D \subset \bigcup_{i=1}^{n}\Psi_i$. As in mobile settings, any data sample $x \in U$ can be encountered where $U$ is the universal set of all possible data, the online mobile inference problem is to identify an optimal subset of $M$, denoted as $M^*$, that maximize the prediction accuracy for $x$. The problem can be discussed in the following three cases of different difficulties: 1) $x \in D$: in this case, $M^*$ is known since $x$ is seen before, \emph{i.e.}, $M^*=\{\mathcal{M}_i: x \in \Psi_i, i \in [1,n] \}$; 2) $x \notin D$ and $x \in \bigcup_{i=1}^{n}\Psi_i$: in this case, $x$ is not seen before and $M^*=\{\mathcal{M}_i: x \in \Psi_i, i \in [1,n] \}$ exists but how to find the $M^*$ is hard; 3) $x \in U-\bigcup_{i=1}^{n}\Psi_i$: in this case, as $x$ is not seen before and $M^*$ does not exist regarding existing $M$, how to make best-effort online prediction for $x$ is challenging. A remedy for this case is to train new models to deal with $x$ and the like in the future.

The main difficulty of the online mobile inference problem lies in \emph{how to determine whether an unseen $x$ belongs to $\Psi_i$ for $i \in [1,n]$}. According to Proposition 1, simply comparing the similarity of semantic attributes between $x$ and $\Gamma_i$ for $i \in [1,n]$ would not work. Another concern is  \emph{how to achieve the best-effort inference accuracy within a specific latency budget even if $M^*$ does not exist}.

\section{Empirical Study}

We first investigate how DNNs behave in mobile settings, taking the typical online object detection as the example task. 

\subsection{Driving Video Datasets}
\label{sec:rdv}
We consider three representative driving video/image datasets as follows:
\begin{itemize}
	\item KITTI~\cite{geiger2012we}: comprises 389 stereo and optical flow image pairs, stereo visual odometry sequences of 39.2 km length, and more than 200k 3D object annotations captured in cluttered scenarios (up to 15 cars and 30 pedestrians are visible per image). For online object detection, KITTI consists of 21 training sequences and 29 test sequences.
	\item BDD100k~\cite{bdd100k}: contains over 100k video clips regarding ten autonomous driving tasks. Clips of 720p and 30fps were collected from more than 50 thousand rides in New York city and San Francisco Bay Area, USA. Each clip lasts for 40 seconds and is associated with semantic attributes such as the scene type (\emph{e.g.}, city, streets, residential areas, and highways), weather condition and the time of the day.
	\item SHD: contains 100 driving video clips of one minute recorded in March 2022 with a 1080p dashcam in Shanghai city, China. Clips were collected from ten typical scenarios, including highway, typical surface roads, and tunnels, at different time in the day. LabelImg \cite{tzutalin2015labelimg} is employed to label objects in all images.
\end{itemize}

\begin{figure}[]
	\centering
	\subfigure[CDF of image brightness]{
		\begin{minipage}[b]{0.22\textwidth}
			\includegraphics[height=2.5cm]{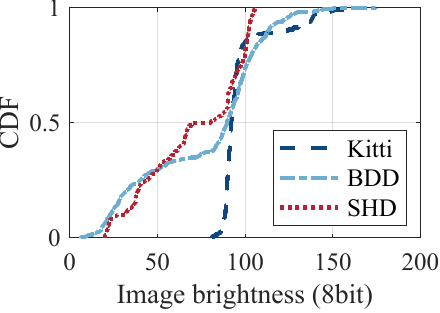}
		\end{minipage}
		\label{fig:brightness}
	}
	\subfigure[CDF of image contrast]{
		\begin{minipage}[b]{0.22\textwidth}
			\includegraphics[height=2.5cm]{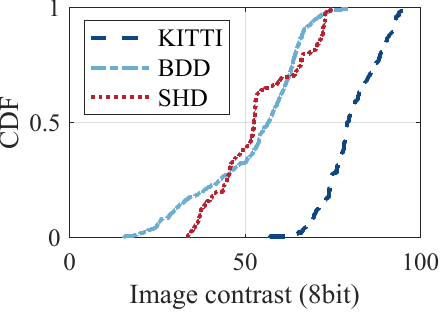}
		\end{minipage}
		\label{fig:contrast}
	}
	\subfigure[CDF of the number of objects]{
		\begin{minipage}[b]{0.22\textwidth}
			\includegraphics[height=2.5cm]{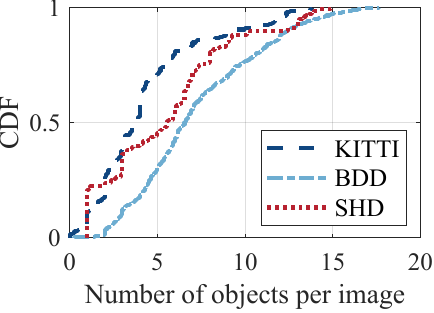}
		\end{minipage}
		\label{fig:number}
	}
	\subfigure[CDF of the ratio of object area]{
		\begin{minipage}[b]{0.22\textwidth}
			\includegraphics[height=2.5cm]{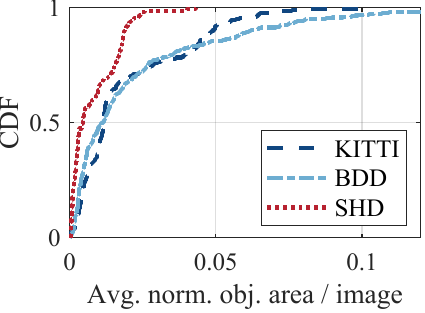}
		\end{minipage}
		\label{fig:area}
	}
	\caption{The dataset of 64 randomly selected driving video clips demonstrates a large diversity in terms of image light conditions and foreground object distributions.}
	\label{fig:diversity}
\end{figure}

We random select 10 video clips from KITTI, 44 clips from BDD100k, and 10 clips from SHD, forming a dataset of 64 video clips containing 16,145 image samples in various scenarios. Figure~\ref{fig:diversity} shows the cumulative distribution functions (CDFs) about foreground objects and illumination condition over all frames in the dataset, demonstrating diverse driving scenarios.
We partition these 64 clips into seen (\emph{i.e.}, involved in model training) and unseen (\emph{i.e.}, not used in model training) categories with a ratio of 9:1. For each seen clip, we further divide frames into training, validation, and testing image sets with a ratio of 6:2:2. 

\subsection{Mobile Inference Accuracy Analysis}

We first train a big object detection model based on YOLOv3, denoted as $\mathcal{M}_{big}$, and a compressed object detection model based on YOLOv3-tiny, denoted as $\mathcal{M}_{0}$, using all available training data of 8,714 images in seen clips. Grid search is used to choose the hyper-parameters during model training. Particularly, $\mathcal{M}_{big}$ incurs 10$\times$  computational cost for inference compared to $\mathcal{M}_{0}$.

\begin{figure*}[]
	\centering
	\subfigure[Boxplot of F1 score, where $\mathcal{M}_{best}$ denotes the best compressed models of each scene]{
		\begin{minipage}[b]{0.30\textwidth}
			\centering
			\includegraphics[height=3cm]{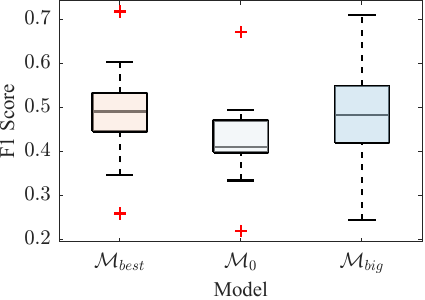}
		\end{minipage}
		\label{fig:shallow_deep}
	}
	\hspace{0.1cm}
	\subfigure[Heatmap of normalized F1 score tested on ten selected scenes defined with semantic attributes]{
		\begin{minipage}[b]{0.31\textwidth}
			\centering
			\includegraphics[height=3cm]{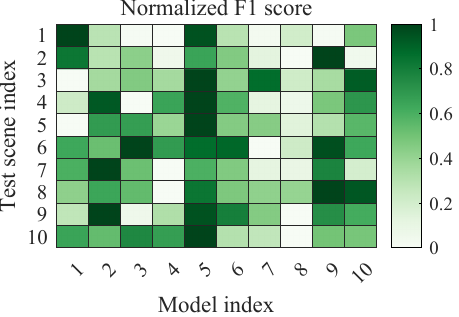}
		\end{minipage}
		\label{fig:video_heatmap}
	}
	\hspace{0.1cm}
	\subfigure[Heatmap of normalized F1 score tested on ten selected scenes defined based on feature similarity]{
		\begin{minipage}[b]{0.31\textwidth}
			\centering
			\includegraphics[height=3cm]{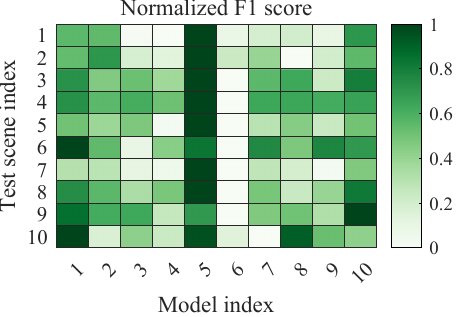}
		\end{minipage}
		\label{fig:feature_heatmap}
	}
	\caption{A best-fit compressed model can achieve comparable prediction accuracy provided by a large DNN, whereas compressed DNNs trained using existing scene partitioning methods fail to perform well on their corresponding training dataset.}
\end{figure*}

Then, we further define ten distinct scenes, denoted as $\Gamma_{i}$ for $i \in [1,10]$, based on the associated semantic attributes of all available training samples, such as \{\emph{good weather, daytime, urban}\}, and for each scene, we train a compressed object detection models, denoted as $\mathcal{M}_{i}$ for $i \in [1,10]$.  	

Figure~\ref{fig:shallow_deep} plots the boxplot of the F1 score obtained with different models on all the 58 testing sets of seen clips. It can be seen that the general compressed model $\mathcal{M}_{0}$ is not the best compressed model. Moreover, we have the following observation:

\begin{observation}
	\emph{Though any single compressed model generally has a lower prediction accuracy than the big model, there exists a compressed model that can achieve comparable accuracy as the big model for each specific scene.}
\end{observation}

From the above observation, it is possible to achieve appealing inference accuracy of the big model at a low cost of a compressed model, if we can identify an appropriate compressed model that best fits the current test sample. An intuitive model selection scheme is to choose the compressed model $\mathcal{M}_{k}$ trained on the dataset $\Gamma_k$ with similar semantic attribute values to the current test sample. This idea assumes that $\Gamma_k \subset \Psi_k$. Figure~\ref{fig:video_heatmap} plots the normalized F1 score for each $\mathcal{M}_{k}$ tested on each $\Gamma_k$. It can be seen that the highest F1 scores do not consistently appear along the diagonal, which makes the intuitive model selection scheme hard to work. Another scheme is to define scenes based on feature similarity, where data samples with similar features are clustered. Given identified scenes, compressed DNNs can be trained and the best DNN can be selected according to the feature similarity between the test samples and existing scenes. Similar model selection results can be seen in Figure~\ref{fig:feature_heatmap}.

\section{Overview of Anole}

\begin{figure*}[t]
	\centering
	\includegraphics[height=9cm]{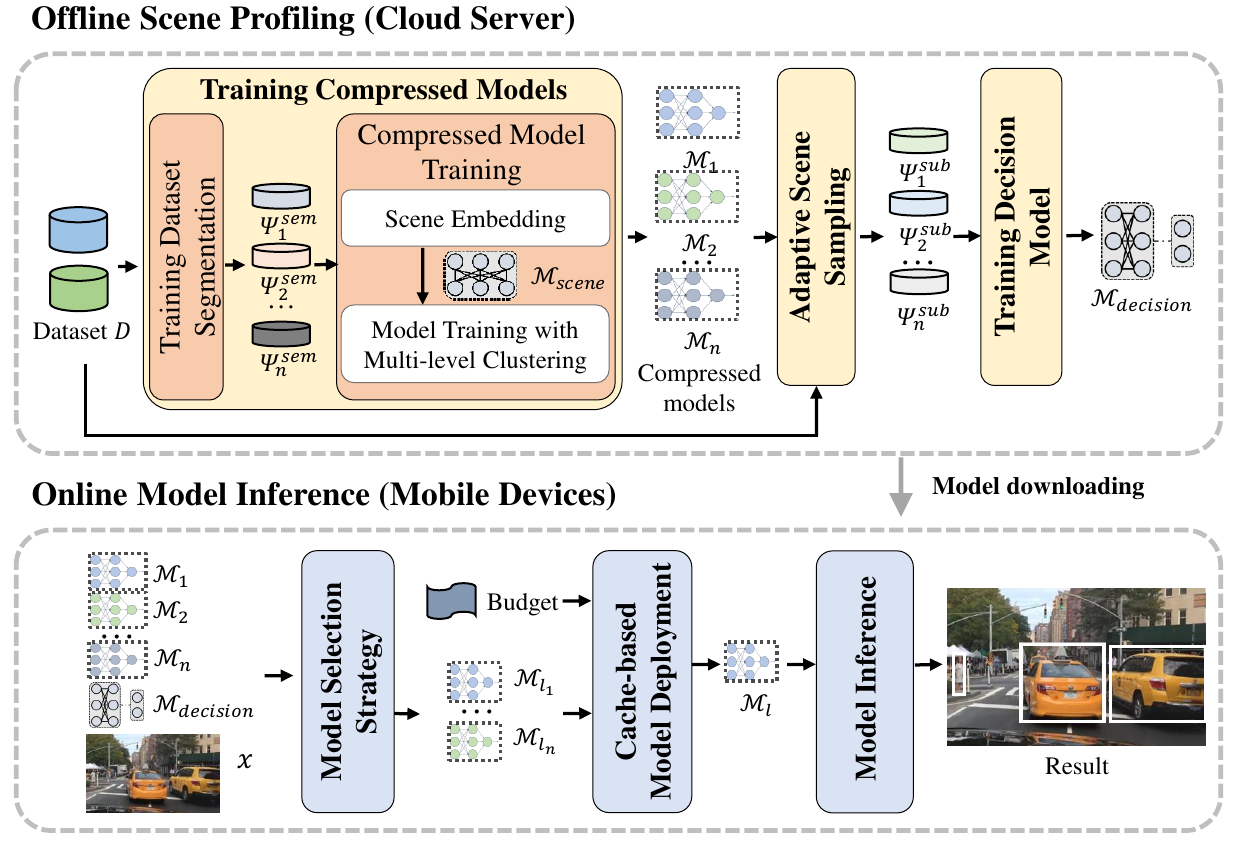}
	\caption{System architecture of Anole, which consists of the offline scene profiling on cloud servers and the online model inference on mobile devices. Communication between both parts is carried out offline.}
	\label{fig:overview}
\end{figure*}

As illustrated in Figure~\ref{fig:overview}, Anole consists of two parts, \emph{i.e.}, offline scene profiling and online model inference.	

\textbf{Offline Scene Profiling (OSP).} OSP is deployed on cloud servers for offline scene partitioning and scene-specific model training, which integrates three components as follows:

\emph{1) Training Compressed Models (TCM):} Given the available labelled dataset $D$, TCM first divides $D$ into appropriate training datasets and train a scene representation model $\mathcal{M}_{scene}$ and a pack of $n$ compressed models $M=\{\mathcal{M}_1, \mathcal{M}_2, \cdots, \mathcal{M}_n\}$; 

\emph{2) Adaptive Scene Sampling (ASS):} As $\{\Psi_1, \Psi_2, \cdots, \Psi_n\}$ are implicit, ASS is to adaptively sample $\{\Psi_1, \Psi_2, \cdots, \Psi_n\}$ based on Thompson sampling from all available dataset $D$ to obtain balanced subsets of $\{\Psi_1, \Psi_2, \cdots, \Psi_n\}$ in $D$, denoted as $\{\Psi_1^{sub}, \Psi_2^{sub}, \cdots, \Psi_n^{sub}\}$, which can be used as labels for decision model training; 

\emph{3) Training Decision Model (TDM):} An end-to-end decision model $\mathcal{M}_{decision}$ is trained using $\{\Psi_1^{sub}, \Psi_2^{sub}, \cdots, \Psi_n^{sub}\}$, which can be used to select suitable compressed models for testing samples.

\textbf{Online Model Inference (OMI).} OMI is deployed on mobile devices for online model inference. Before online inference, pre-trained $\{\mathcal{M}_1, \mathcal{M}_2, \cdots, \mathcal{M}_n\}$ and $\mathcal{M}_{decision}$ need to be downloaded. The core idea of OMI is to compare testing data samples with $\{\Psi_1, \Psi_2, \cdots, \Psi_n\}$ in feature space and select the most suitable compressed models for model inference. To this end, OMI integrates two components:

\emph{1) Model Selection Strategy (MSS):} During online inference, test sample, denoted as $x_{test}$, will be fed to the $\mathcal{M}_{decision}$, which predicts the suitability probability of $\mathcal{M}_i$ for all $i \in [1,n]$ with respect to $x_{test}$. These probabilities are used for ranking models.

\emph{2) Cache-based Model Deployment (CMD):} Given the model ranking, CMD identifies the model with the highest suitability probability in the model cache, denoted as $\mathcal{M}_{test}$, for online inference. If the model with the highest suitability probability is missed, CMD takes the LFU strategy to update models in the cache.

\emph{3) Model Inference (MI):} $\mathcal{M}_{test}$ is applied to $x_{test}$ for conducting local prediction.

\section{Offline Scene Profiling}
\subsection{Training Compressed Models}
\label{subsec:TCM}



\subsubsection{Training Dataset Segmentation}
We first define semantic scenes based on semantic attributes of data.
%
It is non-trivial, however, to manually define appropriate scenes as semantic attributes have different dimensions and different granularities. For example, for driving images, ``urban'' and ``daytime'' are spatial and temporal attributes, respectively, in different dimensions; ``urban'' and ``street'' are spatial attributes but in different granularities. Scenes defined with fine-grained attributes would have insufficient number of samples to train a model whereas scenes defined with coarse-grained attributes would lose the diversity of models. Specifically, we heuristically select fine-grained attributes in each orthogonal dimension to separate data samples into $m$ scenes, denoted as $\{\Gamma_1^{sem}, \Gamma_2^{sem}, \cdots, \Gamma_m^{sem}\}$. For instance, as for driving images, we define semantic scenes according to 120 combinations of attributes in three dimensions, \emph{i.e.}, \{\emph{clear}, \emph{overcast}, \emph{rainy}, \emph{snowy}, \emph{foggy}\} in weather, \{\emph{highway}, \emph{urban}, \emph{residential}, \emph{parking lot}, \emph{tunnel}, \emph{gas station}, \emph{bridge}, \emph{toll booth}\} in location and \{\emph{daytime}, \emph{dawn/dusk}, \emph{night}\} in time\footnote{Note that these scenes are defined at a very fine-grained level, to the extent that they may not have enough samples to train a satisfactory model. They will be clustered further to a moderate granularity for model training.}.

\subsubsection{Compressed Model Training}
\label{sec:cmt}

We employ a training strategy, integrating both semantic similarity and feature similarity of data samples to train diverse compressed models, which consists of the following two steps, as described in Algorithm \ref{Alg:CMT_DM}. 

\textbf{Scene Embedding.} Given semantic scenes $\{\Gamma_1^{sem}, \Gamma_2^{sem}, \cdots, \Gamma_m^{sem}\}$, we train a scene classifier, denoted as $\mathcal{M}_{scene}$, using samples in each $\Gamma_i$ and the index of the scene as label. For each scene dataset $\Gamma_i$ for $i \in [1,m]$, the hidden features on the last layer of $\mathcal{M}_{scene}$, denoted as $H_i$, are used as the embeddings of $\Gamma_i$. 

\textbf{Model Training with Multi-level Clustering.} Instead of training compressed models directly from $\Gamma_i$ for $i \in [1,m]$, we further consider the feature similarity of data samples by clustering embeddings in all $H_i$ and train compressed models on obtained clusters. 
Specifically, two versions of multi-level clustering algorithms are proposed in Anole, \emph{i.e.}, Compressed Model Training with Distinct Samples among Models (CMT-DM,  described in Algorithm \ref{Alg:CMT_DM}), and Compressed Model Training with Shared Samples among Models (CMT-SM, described in Algorithm \ref{Alg:CMT_SM}).

\begin{algorithm}[]
	\KwIn{Semantic-defined scenes $\Gamma_i^{sem}$ for $i \in [1,m]$, preset number $n$ of compressed models to be trained, threshold $\delta$ at which the model performance meets the required criteria, embeddings of all scenes $H_i$ for $i \in [1,m]$.}
	\KwOut{Compressed models specific for scenes $M^{rep} = \{\mathcal{M}_1, \mathcal{M}_2, \cdots, \mathcal{M}_n\}$.}
	\BlankLine
	\caption{CMT-DM Algorithm} \label{Alg:CMT_DM}
	\tcp{\footnotesize Model training with multi-level clustering.}
	Compressed model repository $M^{rep} \gets \{\}$\;
	clustering number $k \gets 2$\;
	\While{$|M^{rep}| < n$}{
		Cluster on $\{H_1, H_2, \cdots, H_m\}$ with clustering number $k$\;
		Train $k$ compressed models $\mathcal{M}^k_j$ for $j \in [1,k]$\;
		\For{each $\mathcal{M}^k_j$}
		{	$p_j \gets$ evaluation performance of $\mathcal{M}^k_j$ on its vaildation set\;
			\If{$p_j > \delta$}
			{
				$\mathcal{M}_{|M^{rep}| + 1} \gets \mathcal{M}^k_j$\;
				$M^{rep}$.append($\mathcal{M}_{|M^{rep}| + 1}$)\;
			}
		}
		$k \gets k+1$\;
	}
	\Return{$M^{rep} = \{\mathcal{M}_1, \mathcal{M}_2, \cdots, \mathcal{M}_n\}$}
\end{algorithm}

The core idea of the CMT-DM algorithm is to obtain the clusters with different levels of similarity.
To this end, we conduct multiple $k$-means \cite{lloyd1982least} clustering with $k$ varying from 2 over embeddings in all $H_i$ for $i \in [1,m]$. For each $k$, all embeddings can be divided into $k$ clusters, denoted as $C^k_j$ for $j \in [1,k]$ (Line 4). We train a compressed model, denoted as $\mathcal{M}^k_j$, on each clustered scene corresponding to $C^k_j$ for $j \in [1,k]$ (Line 5) and validate its performance. If the performance of $\mathcal{M}^k_j$ exceeds a threshold $\delta$, $\mathcal{M}^k_j$ is added to the compressed model repository (Line 6 - Line 10). This procedure repeats until a set of $n$ compressed models $\{\mathcal{M}_1, \mathcal{M}_2, \cdots, \mathcal{M}_n\}$ are derived, where $n$ denotes a preset number for compressed models to be trained.

\begin{algorithm}[]
	\KwIn{Semantic-defined scenes $\Gamma_i^{sem}$ for $i \in [1,m]$, preset number $n$ of compressed models to be trained, threshold $\delta$ at which the model performance meets the required criteria, embeddings of all scenes $H_i$ for $i \in [1,m]$, initial clustering number $K$.}
	\KwOut{Compressed models specific for scenes $M^{rep} = \{\mathcal{M}_1, \mathcal{M}_2, \cdots, \mathcal{M}_n\}$.}
	\BlankLine
	\caption{CMT-SM Algorithm} \label{Alg:CMT_SM}
	\tcp{\footnotesize Model training with multi-level clustering.}
	Compressed model repository $M^{rep} \gets \{\}$\;
	$\bm{C} = \{C_1, C_2, \cdots, C_K\} \gets $ clusters on $\{H_1, H_2, \cdots, H_m\}$ with clustering number $K$\;
	\While{$|M^{rep}| < n$}{
		Train $K$ compressed models $\mathcal{M}_j$ for $j \in [1,K]$ on clusters $\{C_1, C_2, \cdots, C_K\}$\;
		\tcp{\footnotesize performance evaluation for each candidate compressed models.}
		\For{each $\mathcal{M}^k_j$}
		{	$p_j \gets$ evaluation performance of $\mathcal{M}^k_j$ on its vaildation set\;
			\If{$p_j > \delta$}
			{
				$\mathcal{M}_{|M^{rep}| + 1} \gets \mathcal{M}^k_j$\;
				$M^{rep}$.append($\mathcal{M}_{|M^{rep}| + 1}$)\;
			}
		}
		$\bm{C}^s = \{C_1^s, C_2^s, \cdots, C_K^s\} \gets $ ascending sorted clusters according to their performance\;
		\tcp{\footnotesize clusters combination.}
		\For{each $C_{k_1}^s$ for $k_1 \in [1, K]$}{
			\For{each $C_{k_2}^s$ for $k_2 \in [2, K]$}{
				$p_{k_1,k_2} \gets$ performance of the model $\mathcal{M}_{k_1, k_2}$ trained on $C_{k_1,k_2}^s = C_{k_1}^s \cup C_{k_2}^s$\;
				\If{$p_{k_1,k_2} > p_{k_1}$ and $p_{k_1,k_2} > p_{k_2}$}{
					$\bm{C} \gets \bm{C} \cup \{C_{k_1,k_2}^s\} \setminus \{C_{k_1}^s, C_{k_2}^s\}$\;
				}
				\ElseIf{$p_{k_1,k_2} > p_{k_1}$ and $p_{k_1,k_2} \le p_{k_2}$}{
					$\bm{C} \gets \bm{C} \cup \{C_{k_1,k_2}^s\} \setminus \{C_{k_1}^s\}$\;
				}
				\ElseIf{$p_{k_1,k_2} \le p_{k_1}$ and $p_{k_1,k_2} > p_{k_2}$}{
					$\bm{C} \gets \bm{C} \cup \{C_{k_1,k_2}^s\} \setminus \{C_{k_2}^s\}$\;
				}
			}
		}
	}
	\Return{$M^{rep} = \{\mathcal{M}_1, \mathcal{M}_2, \cdots, \mathcal{M}_n\}$}
\end{algorithm}

The core idea of the CMT-SM algorithm is to first cluster scenes at a fine-grained level and then to try to combine the clusters with bad model training performance by pairwise combination trail.
To this end, we first conduct $k$-means clustering with a preset clustering number $K$ over embeddings in all $H_i$ for $i \in [1,m]$. All embeddings can therefore be divided into $K$ clusters, denoted as $\bm{C} = \{C_1, C_2, \cdots, C_K\}$, to get a fine-grained partitioning (Line 2). We train a model for each clusters on $\bm{C}$ and the models whose performance exceeds the threshold $\delta$ will be added to the compressed model repository (Line 5 - Line 9). Then, we sort $\bm{C} = \{C_1, C_2, \cdots, C_K\}$ in ascending validated performance order to obtain $\bm{C}^s = \{C_1^s, C_2^s, \cdots, C_K^s\}$ for further combination of clusters (Line 10). Next, we start from the two clusters with the worst training performance, \emph{i.e.}, $C_1^s$ and $C_2^s$, attempting to combine them pairwise, and verify whether the training performance, denoted as $p_{k_1,k_2}$ of combined cluster, denoted as $C_{k_1,k_2}^s = C_{k_1}^s \cup C_{k_2}^s$, is higher than that of $C_1^s$ and $C_2^s$, denoted as $p_{k_1}$ and $p_{k_2}$, respectively (Line 11 - Line 19). If $p_{k_1,k_2}$ is larger than both $p_{k_1}$ and $p_{k_2}$, $C_1^s$ and $C_2^s$ will be removed from the clusters set $\bm{C}$ and $C_{k_1,k_2}^s$ will be added; If only $p_{k_1,k_2}$ is larger than only one of $p_{k_1}$ or $p_{k_2}$, only the less one ($p_{k_1}$ or $p_{k_2}$ which is less) will be removed from $\bm{C}$ and $C_{k_1,k_2}^s$ will be added. This procedure repeats until a set of $n$ compressed models are derived.

\subsection{Adaptive Scene Sampling}

\begin{figure}[]
	\centering
	\subfigure[Random sampling]{
		\begin{minipage}[b]{0.22\textwidth}
			\centering
			\includegraphics[height=2.8cm]{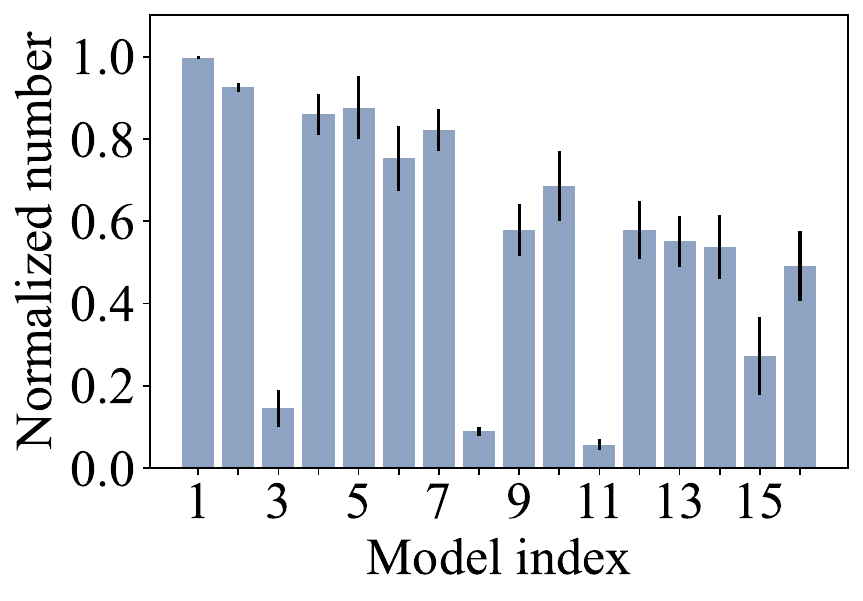}
		\end{minipage}
		\label{fig:label_unbalanced}
	}
	\subfigure[Adaptive sampling]{
		\begin{minipage}[b]{0.22\textwidth}
			\centering
			\includegraphics[height=2.8cm]{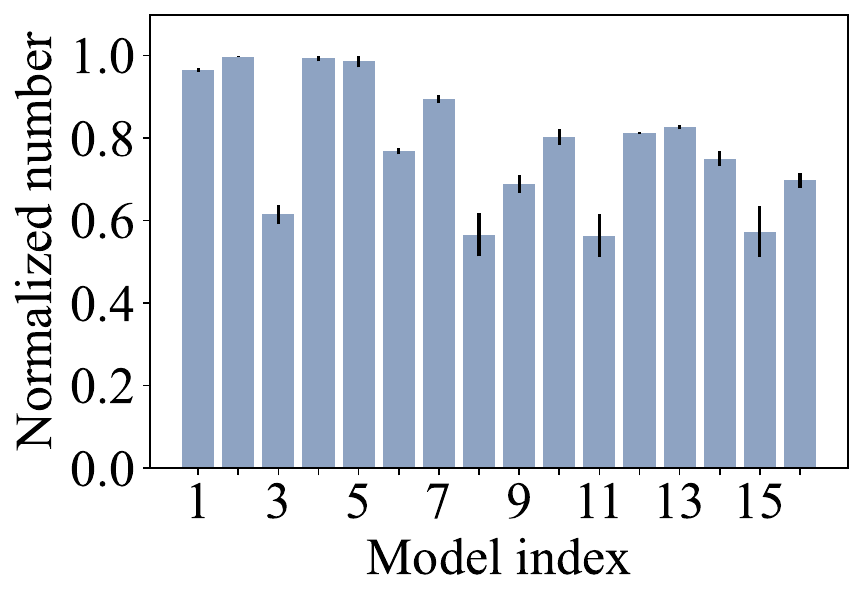}
		\end{minipage}
		\label{fig:label_balanced}
	}
	\caption{(a) An example of compressed models being unevenly sampled with random sampling; (b) our adaptive sampling algorithm can mitigate the unbalanced sampling problem.}
	\label{fig:label}
\end{figure}

To obtain $\{\Psi_1^{sub}, \Psi_2^{sub}, \cdots, \Psi_n^{sub}\}$, a straightforward idea is to randomly pick a number of samples $X$ from $D$ and test $\mathcal{M}_i$ for $i \in [1, n]$. If a $\mathcal{M}_i$ can achieve satisfactory prediction accuracy on sample $x \in X$, $x$ belongs to $\Psi_i^{sub}$. As $\{\Psi_1, \Psi_2, \cdots, \Psi_n\}$ may be biased in $D$, such random sampling algorithm also generates unbalanced $\{\Psi_1^{sub}, \Psi_2^{sub}, \cdots, \Psi_n^{sub}\}$. 
To solve the unbalanced sampling problem, however, is not intuitive, because of Proposition \ref{proposition1}. Proposition \ref{proposition1} holds that we can not know a sample belongs to which distribution from all the distributions those models can characterize (\emph{i.e.}, $\{\Psi_1, \Psi_2, \cdots, \Psi_n\}$) without high computational cost experiments. In order to obtain a balanced  $\{\Psi_1^{sub}, \Psi_2^{sub}, \cdots, \Psi_n^{sub}\}$ at a low computation cost, we design an adaptive sampling algorithm based on Thompson sampling \cite{thompson}.

Specifically, in the $k$-th sampling round for $k \in \mathbb{N}$, we first examine if the training set $\Gamma_i$ of $\mathcal{M}_i$ for $i \in [1,n]$ has been well sampled by checking $|S_i| > \frac{\log (1-\theta^{\frac{1}{|\Gamma_i|}})}{\log(1-\frac{1}{|\Gamma_i|})}$,
where $S_i$ is the set of samples sampled from $\Gamma_i$; $\theta$ is the confidence of being well sampled; and $| \cdot |$ is the number of elements in a set. 

Then, for each training set $\Gamma_i$ that has not been well sampled, we estimate a sampling probability $p_i^k$ based on a Beta distribution $Beta(\alpha_i^{k-1}, \beta_i^{k-1})$, where $\alpha_i^{k-1}$ and $\beta_i^{k-1}$ are the two parameters of the Beta distribution of $\Gamma_i$, updated in the previous round. 
As a result, the training set $\Gamma_k$ with the highest sampling probability will be sampled.

Finally, all $Beta(\alpha_i^k, \beta_i^k)$ will be updated as follow:
\begin{equation*}
	Beta(\alpha_i^k, \beta_i^k)=
	\begin{cases}
		Beta(\alpha_i^{k-1}+ 1, \beta_i^{k-1}), & \text{if $\Gamma_i$ is sampled;}\\
		Beta(\alpha_i^{k-1}, \beta_i^{k-1} + 1), & \text{otherwise.}
	\end{cases}
\end{equation*}
This procedure repeats until a specific number of $\kappa$ samples are collected. Figure \ref{fig:label} shows the normalized $|S_i|$ for all the $\mathcal{M}_i$ for $i \in [1,n]$ where $n=16$, using the random sampling algorithm and our adaptive sampling algorithm, respectively. It can be seen that our adaptive sampling algorithm can effectively mitigate the unbalanced sampling problem.

\subsection{Training Decision Model}
\label{sec:tdm}





Given the sampling results $\{\Psi_1^{sub}, \Psi_2^{sub}, \cdots, \Psi_n^{sub}\}$, we train an end-to-end decision model $\mathcal{M}_{decision}$ to effectively represent and distinguish $\{\Psi_1^{sub}, \Psi_2^{sub}, \cdots, \Psi_n^{sub}\}$ by employing a parameter-frozen scene representation network $\mathcal{M}_{scene}$ and neural-network-based classifier.

Specifically, we use $\mathcal{M}_{scene}$ as a backbone neural network to obtain scene representation, denoted as $h^s_i$, for every data sample $x_i \in \Psi_i^{sub}, i \in [1,n]$. In this way, $h^s_i$ will retain the scene-related information. 
The model decision here can be formulated as a multi-class classification problem. 
The label of $x$ for decision model training is a vector, referred to as a \emph{model allocation vector} $v^x = \{v_i^x, i \in [1,n]\}$, where the $i$-th element $v_i^x$, denotes whether $x \in \Psi_i^{sub}$.
The cross entropy loss function~\cite{de2005tutorial} is used for training the decision model. Note that during the training of decision model $\mathcal{M}_{decision}$, the parameter of  $\mathcal{M}_{scene}$ is frozen to improve training efficiency and enhance the generalization of $\mathcal{M}_{decision}$ \cite{Li2022}.

\section{Online Model Inference}
\subsection{Model Selection Strategy}

Given the set of pretrained models $\{\mathcal{M}_1, \mathcal{M}_2, \cdots, \mathcal{M}_n\}$ and decision model $\mathcal{M}_{decision}$ downloaded from a cloud server, a mobile device needs to select most suitable compressed models for online inference. 
Specifically, it utilizes $\mathcal{M}_{decision}$ to output the model allocation vector $v^x$ for a testing sample $x$, \emph{i.e.}, $v^x=\mathcal{M}_{decision}(x)$, where the $i$-th element $v_i$ indicates the suitability probability that model $\mathcal{M}_i$ is suitable for $x$. Therefore, we can rank all compressed models according to their suitability probablities for $x$ using $v^x$. It should be noted that for the uncertainty of scenario duration, model selection should be conducted on every testing sample, taking into account the fast-changing data distributions in the perspective of compressed models.





\subsection{Cache-based Model Deployment}

With the model allocation vector $v^x=\mathcal{M}_{decision}(x)$, compressed models can be dynamically ranked. Due to the restricted amount of memory on a mobile device, not all models may be pre-loaded into memory. To deal with this issue, we investigate the best-effort model deployment strategy.

\begin{figure}[]
	\centering
	\subfigure[Average inference latency]{
		\begin{minipage}[b]{0.22\textwidth}
			\centering
			\includegraphics[height=2.8cm]{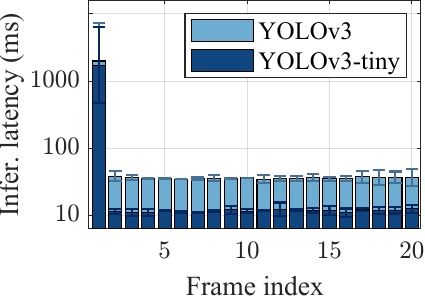}
		\end{minipage}
		\label{fig:multi-latency}
	}
	\subfigure[Utility of compressed models]{
		\begin{minipage}[b]{0.22\textwidth}
			\centering
			\includegraphics[height=2.8cm]{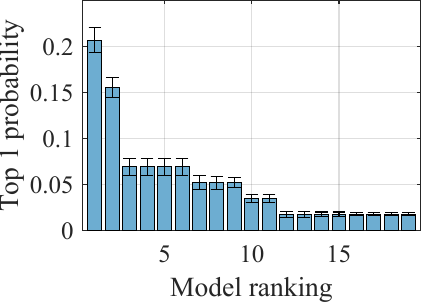}
		\end{minipage}
		\label{fig:top1}
	}
	\caption{(a) Average latency of model inference on consecutive frames over all test clips; (b) the probability of being the top one model, following a long-tailed distribution.}
	\label{fig:multi_model}
\end{figure}

We examine the inference latency of detecting objects on five driving video clips, using two DNN models of different size, \emph{i.e.}, YOLOv3 (237MB) and YOLOv3-tiny (33.8MB), on a Nvidia Jetson TX2 NX (ARM A57 CPU, Nvidia Pascal GPU with 4GB memory, 32GB flash). Figure~\ref{fig:multi-latency} plots the average inference latency of the first twenty frames over all clips. For both models, a huge delay occurs when processing the first frame. This is mainly attributed to the I/O operation for model loading and other initialization required by the deep learning framework such as Pytorch. Therefore, it is preferred to preload as many models as possible. 


Given a limited video memory budget, it is tricky to pre-load best models in memory. We examine the utility of 19 YOLOv3-tiny compressed models obtained according to the algorithm stated in $\S$\ref{subsec:TCM} (see $\S$\ref{subsubsection:implementation} for more details) when conducting object detection on the five driving video clips. Figure~\ref{fig:top1} depicts the ratio of being the top one model over all clips for all compressed models. It can be seen that the probability of being the best model follows a power-law distribution. This observation suggests that high-level inference performance can be sustained by deploying only a small number of supreme models. Inspired by this observation, we adopt a Least Frequently Used (LFU) strategy \cite{silberschatz2018operating} to update models in GPU memory. In the occasion of a model miss, the model with the highest suitability probability in GPU memroy will be used for inference.

\section{Evaluation}
\label{sec:eval}

\subsection{Methodology}
\subsubsection{Tasks and Datasets} 
\begin{figure}[t]
	\centering
	\includegraphics[height=2.5cm]{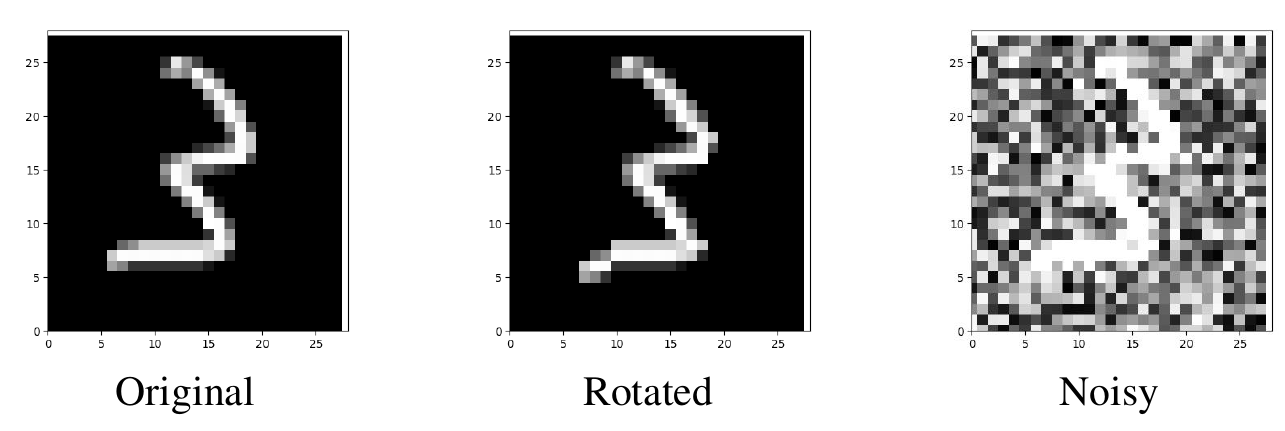}
	\caption{Example images in the original and two constructed MNIST datasets.}
	\label{fig:mnist}
\end{figure}

We evaluate Anole on two mobile inference tasks, \emph{i.e.}, handwritten digit recognition (HDR), coke grades prediction on mobile agents (CGP) and vehicle detection on driving videos (VD), based on the following datasets and real-world experiments.
\begin{itemize}
	\item \textbf{Handwritten Digit Datasets}: The MNIST dataset \cite{lecun1998gradient} contains 70,000 black and white images of handwritten digits. The images are normalized to a 28$\times$28 pixel bounding box and anti-aliased. As illustrated in Figure \ref{fig:mnist}, we also construct two datasets, \emph{i.e.}, rotated MNIST and noisy MNIST, by applying random rotation ranging from $-45^{\circ}$ to $45^{\circ}$ and random gaussian noise with zero mean and a variance of 1 to the original MNIST images, respectively. We mix all three datasets and partition all images into seen and unseen categories with a ratio of 3:2. We further divide seen images into training, validation, and testing sets with a ratio of 2:1:1. 
	\item \textbf{Coking Production Decision Making Dataset}: The Coke Production Dataset (CPD) is collected from a leading international cloud service provider. It is to predict coke grades for coke production operations where decisions with the highest grades will be taken. A total of 200 coke production experiments collected from July, 2020 to March, 2021 are collected in the CPD dataset. Each sample includes structured features such as the ratio of different coal raw materials and the quality of the refined coke is labeled for prediction. Such a prediction can be used to optimize the decision of industrial production. The samples in the CPD dataset vary from multiple factors such as raw material loactions and production specifications owing to the mobility among production in different batches. We divide the samples into training, validation and testing sets with a ratio of 3:1:1.
	\item \textbf{Driving Video Datasets}: We use the established driving video dataset comprising 64 clips randomly selected from KITTI, BDD100k and SHD, with seen and unseen data divided as introduced in $\S$\ref{sec:rdv}.  
\end{itemize}

\begin{table}[]
	\caption{Anole is implemented on three different types of mobile devices with distinct hardware configurations.}
	\label{tab:hardware}
	\centering
	\scalebox{1}{
		\begin{tabular}{c c c c c c}
			\toprule
			Platform               & CPU & GPU & GPU Mem. & Flash/Disk \\ \hline
			Jetson Nano   & ARM A57  & Maxwell  & 2GB    & 32GB \\ 
			Jetson TX2 & ARM A57  & Pascal & 4GB   & 32GB   \\ 
			Laptop & i7-10750H  & RTX 2070    & 8GB    & 1T   \\ 
			\bottomrule
		\end{tabular}
	}
\end{table}

\subsubsection{Implementation} 
\label{subsubsection:implementation}

We implement the offline scene profiling on a server equipped with 128GB RAM and 4 Nvidia 2080 Ti GPUs, running a Linux distribution. We implement online model inference on three typical mobile devices, \emph{i.e.}, a Nvidia Jetson Nano, a Nvidia Jetson TX2 NX and a Windows laptop. Pytorch is employed as the inference engine and TensorRT \cite{vanholder2016efficient} is used for the run-time acceleration on both Jetson devices, running a Linux distribution.
OpenCV is compiled on CPU for balancing the usage of CPU and GPU. 
The hardware configurations are shown in Table~\ref{tab:hardware}. 
For the HDR task, because of small image size, we direct conduct multi-level clustering on image pixels without training the $\mathcal{M}_{scene}$ and use a multi-layer perceptron (MLP) \cite{kruse2022multi} with two layers to train the $\mathcal{M}_{decision}$ for its light-weight characteristic. Compressed models for image classification are trained based on LeNet \cite{lecun1998gradient}.
For the CGP task, we directly use the structured features without training the $\mathcal{M}_{scene}$ and use a decision tree (DT) model \cite{breiman2017classification} based on Scikit-Learn \cite{scikit-learn} as $\mathcal{M}_{decision}$ for its superior performance and interpretability. Compressed models for coke grade prediction are trained based on Linear Regression (LinearR) \cite{rifkin2007notes}.
For the VD task, ResNet18 \cite{he2016deep} and a MLP of two layers are used to train the $\mathcal{M}_{scene}$ and the $\mathcal{M}_{decision}$, respectively. 
Compressed models for object detection are fine-tuned on YOLOv3-tiny \cite{adarsh2020yolo} pre-trained on the COCO~\cite{lin2014microsoft} dataset. Details of all deployed models are listed in Table~\ref{tab:models}.
Compressed models of HDR and VD tasks are trained with Algorithm \ref{Alg:CMT_DM} consodering the heavy training overhead of DNN model and compressed models of CGD tasks are trained with Algorithm \ref{Alg:CMT_SM} for a better performance. A total of 25, 17 and 19 compressed models are trained for the HDR, CGD and VD tasks, respectively, to provide compressed models for inference in all possible scenes.


\begin{table}[]
	\caption{Details of deployed models, where FLOPS of the deep model YOLOv3 is $10\times$ bigger than YOLOv3-tiny and Resnet18.}
	\label{tab:models}
	\centering
	\begin{tabular}{ccccc}
		\toprule
		Task                 & Model        & Role                     & FLOPS    & Weights \\ \hline \hline
		\multirow{4}{*}{HDR} & LeNet        & Compress model           & 31 M     & 234 KB  \\  
		& MLP          & $\mathcal{M}_{decision}$ & 3.6 M    & 935 KB  \\ 
		& Resnet18     & Deep model               & 4.69 Bn  & 44 MB   \\  
		& Resnet50     & Deep model               & 9.74 Bn  & 87 MB   \\ 
		\hline
		\multirow{2}{*}{CGP} & LinearR      & Compress model           &   0.2 K       &   0.1 KB      \\ 
		& DecisionTree & $\mathcal{M}_{decision}$ &  0.1 K       &  1.5 KB       \\\hline
		\multirow{4}{*}{VD}  & YOLOv3-tiny  & Compress model           & 5.56 Bn  & 34 MB   \\ 
		& Resnet18     & $\mathcal{M}_{scene}$    & 4.69 Bn  & 44 MB   \\  
		& MLP          & $\mathcal{M}_{decision}$ & 3.6 M    & 935 KB  \\  
		& YOLOv3       & Deep model               & 65.86 Bn & 237 MB  \\  \bottomrule
	\end{tabular}
\end{table}

\subsubsection{Candidate Methods} 
We compare Anole with the following candidate methods:
\begin{itemize}
	\item \textbf{Single Deep Model (SDM)~\cite{adarsh2020yolo}:} One single deep model is trained with all training samples for online inference. For the HDR task, a ResNet18 and a ResNet50 are respectively trained for their supreme image classification performance. For the VD task, a fully-fledged YOLOv3 is trained. For the CGP task, no deep models are considered because there is no large model good enough so far for this task.
	\item \textbf{Single Shallow Model (SSM)~\cite{redmon2018yolov3}:} One single compressed model is trained with all training samples for online inference. For the HDR task, a LeNet is trained. For the VD task, a YOLOv3-tiny is trained.
	\item \textbf{Clustering-based Domain Generalization (CDG)~\cite{zheng2019metadata}:} Compressed models are trained on domains defined by clustering training data samples in the feature space. During online prediction, the compressed model trained on the cluster which has the closest mean compared with the feature of the test sample is selected for use.
	\item \textbf{Dataset-based Multiple Models (DMM):} One separate compressed model is trained on each training dataset, \emph{i.e.}, the original, rotated, and noisy MNIST for the HDR task, the KITTI, BDD100k, and SHD for the VD task, the human-defined subsets for CGP task. During online prediction, the compressed model corresponding to the same dataset as the test sample is selected for use.
\end{itemize} 

\subsubsection{Metrics}
We evaluate the performance of all candidate methods with respect to inference accuracy and latency. For the HDR task, accuracy is defined as the proportion of correctly predicted samples to the total number of test samples. 
For the CGD task, we use Error Rate (ER), defined as $\text{ER} = \frac{\sum_{n=1}^{N} |\hat{y}_n - y_n|}{\sum_{n=1}^{N} y_n}$, Mean Absolute Error (MAE), defined as $\text{MAE} = \frac{\sum_{i=1}^N |\hat{y}_{i} - y_{i}|}{N}$, Mean Squared Error (MSE), defined as $\text{MSE} = \frac{\sum_{i=1}^N (\hat{y}_{i} - y_{i})^2}{N_j}$ and Pearson Correlation Coefficient (PCCS), defined as $\text{PCCS} = \frac{\sum_{i=1}^{N} (\hat{y}_{i} - \hat{\mu})(y_{i} - \mu)}{\sqrt{\sum_{i=1}^{N} (\hat{y}_{i} - \hat{\mu})^2}\sqrt{\sum_{i=1}^{N}(y_{i} - \mu)^2}}$ for evaluation, where $N$ denotes the number of all testing data; $\hat{y}_n$ and $y_n$ are the estimation and the ground truth of the $n$-th sample; $\hat{\mu}^j$ and $\mu^j$ are the mean of $\hat{y}_n$ and $y_n$, respectively.
For the VD task, we use F1 score, defined as $\text{F1} = \frac{2 \cdot p \cdot r}{p + r}$, where $p$ and $r$ denote the precision and recall of detection, respectively. We also consider the end-to-end delay, \emph{i.e.}, the time duration from receiving a test sample to obtaining the corresponding inference result.

\subsection{Effect of Scene Profiling Models}
\subsubsection{Scene Encoder $\mathcal{M}_{scene}$}
\begin{figure}[]
	\centering
	\subfigure[$\mathcal{M}_{scene}$ for VD]{
		\begin{minipage}[t]{0.145\textwidth}
			\includegraphics[width=\textwidth]{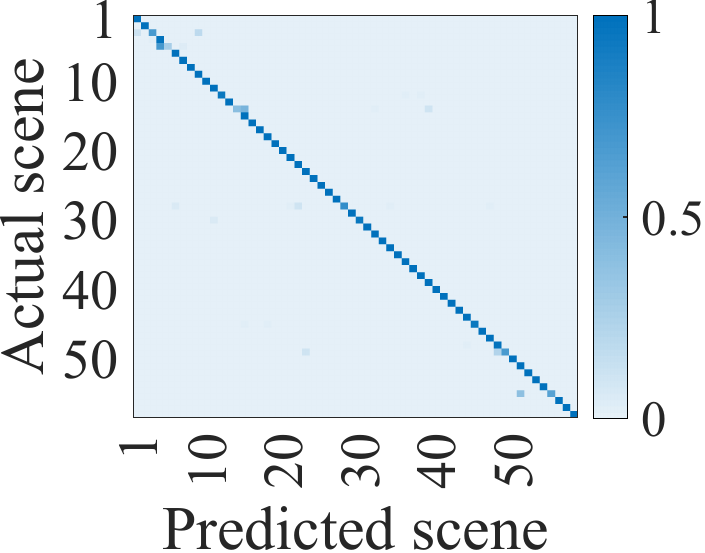}
		\end{minipage}
		\label{fig:model_scene}
	}
	\subfigure[$\mathcal{M}_{decision}$ for HDR]{
		\begin{minipage}[t]{0.145\textwidth}
			\includegraphics[width=\textwidth]{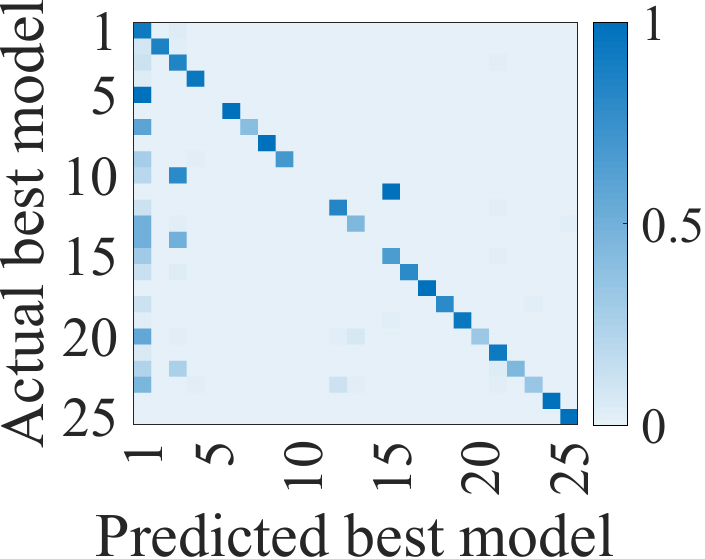}
		\end{minipage}
		\label{fig:model_decision_hdr}
	}
	\subfigure[$\mathcal{M}_{decision}$ for VD]{
		\begin{minipage}[t]{0.145\textwidth}
			\includegraphics[width=\textwidth]{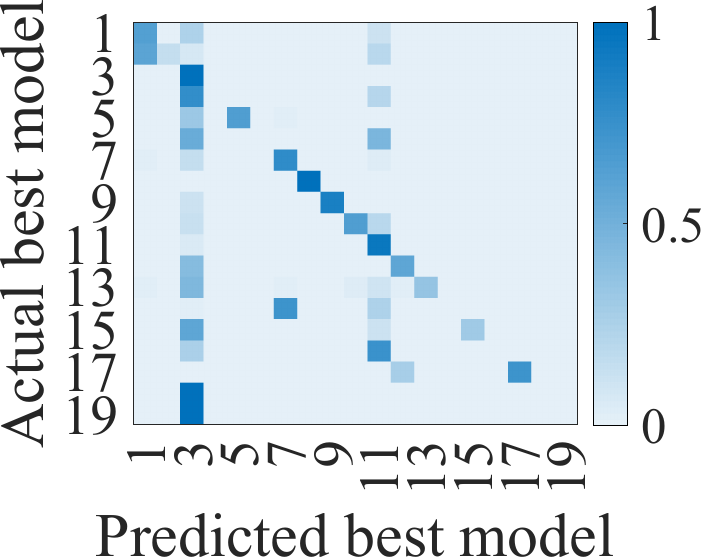}
		\end{minipage}
		\label{fig:model_decision}
	}
	\caption{Confusion matrices of scene profiling models, showing high accuracy for scene encoding and model decision, respectively.}
	\label{fig:selection}
\end{figure}

For the VD task, we test $\mathcal{M}_{scene}$ on classifying scenes on the validation set of seen scenes. Scenes are defined based on the multi-level clustering results.
Figure~\ref{fig:model_scene} shows the scene classification confusion matrix of scene encoder $\mathcal{M}_{scene}$ on the validation set. It can be seen that $\mathcal{M}_{scene}$ works well among almost all scenes. There also exist some exceptional scenes that are confusing to $\mathcal{M}_{scene}$. We merge similar scenes in the feature space before training compressed models. 

\subsubsection{Decision Model $\mathcal{M}_{decision}$}
We evaluate the ability of $\mathcal{M}_{decision}$ in selecting the top-one model on the validation set of seen data.
Figure~\ref{fig:model_decision_hdr} and Figure~\ref{fig:model_decision} show the confusion matrix of the $\mathcal{M}_{decision}$ models predicting best models versus true best models for the HDR and VD tasks, respectively.
It can be seen that $\mathcal{M}_{decision}$ have basic model selection ability. This is because the decision of model selection is based on the well-trained $\mathcal{M}_{scene}$, with one scene corresponding to a group of suitable models. We can also see that $\mathcal{M}_{decision}$ may make mistakes on some models,. This is because the top one model may not be significantly better than other models.

\subsection{Effect of Cache-based Model Update Strategy}
To effectively evaluate the effect of our cache-based model update strategy, we synthesize six fast-changing video clips, denoted as T1-T6, for the VD task. Specifically, for each synthesized video clip, we randomly select 5 clips from the 64 clips in the dataset. For each selected clip, we randomly cut a video segment of 100 frames (from the testing set for a seen clip) and then splice all video segments, resulting a synthesized video clip of 500 frames. We then conduct model inference using Anole on T1-T6.

\begin{figure}[]
	\centering
	\subfigure[Scene duration distributions on synthesized video clips]{
		\begin{minipage}[b]{0.22\textwidth}
			\includegraphics[width=0.9\textwidth]{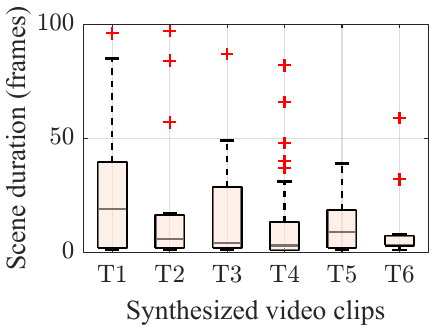}
		\end{minipage}
		\label{fig:during}
	}
	\subfigure[Cache miss rate and F1 score obtained with varying cache size]{
		\begin{minipage}[b]{0.22\textwidth}
			\includegraphics[width=1\textwidth]{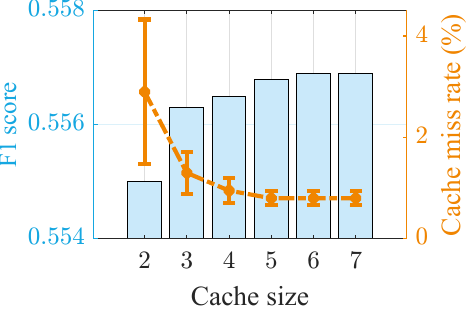}
		\end{minipage}
		\label{fig:cache}
	}
	\caption{(a) Boxplot of scene duration, measured as the number of frames without model switching; (b) cache miss rate and F1 score as functions of varying cache sizes.}
	\label{fig:trace}
\end{figure}

\subsubsection{Scene Duration} Figure~\ref{fig:during} plots the boxplot of scene duration measured as the number of frames without model switching on all six synthesized video clips. It can be seen that scenes change rapidly in the perspective of $\mathcal{M}_{decision}$, with over 80\% of scenes lasting fewer than 40 frames and the mean scene duration less than 20 frames. 

\subsubsection{Cache Miss Rate} Figure~\ref{fig:cache} depicts the cache miss rate and the F1 score as functions of cache size in the unit of compressed model size. It can be seen that a cache capable of loading up to 5 models can sustain a low cache miss rate and a stable inference accuracy. This observation aligns with the observation of the long-tail model utility distribution as shown in Figure~\ref{fig:top1}. It is also observed that the inference accuracy remains satisfactory even for a cache size of 2 models, demonstrating the feasibility of Anole on devices with extremely limited GPU memory.

\begin{table}[]
	\centering
	\caption{Inference accuracy of candidate methods achieved in cross-scene and new-scene scenarios on the HDR task.}
	\label{tab:mnist}
	\scalebox{1}{
		\begin{tabular}{ccccccc}
			\toprule
			\multirow{2}{*}{Method} & \multirow{2}{*}{Model} & \multirow{2}{*}{\begin{tabular}[c]{@{}c@{}}GPU Mem.\\ (MB)\end{tabular}} & \multicolumn{2}{c}{Cross-scene}    & \multicolumn{2}{c}{New-scene}      \\ 
			&         &          & \multicolumn{1}{c}{Noisy} & Rota. & \multicolumn{1}{c}{Noisy} & Rota. \\ \hline
			\multirow{2}{*}{SDM}    & Res.18             & 45.27     & \multicolumn{1}{c}{\color{blue}0.773}      &  \color{blue}0.828      & \multicolumn{1}{c}{\color{blue}0.777}      &   \color{blue}0.826     \\ 
			& Res.50   & 98.31      & \multicolumn{1}{c}{0.704}      &  0.810      & \multicolumn{1}{c}{0.702}      &   0.808     \\ 
			SSM    & LeNet     & 0.18      & \multicolumn{1}{c}{0.755} & 0.751  & \multicolumn{1}{c}{0.754} & 0.752  \\ 
			CDG   & LeNet    & 0.18   & \multicolumn{1}{c}{0.099}      &  0.721      & \multicolumn{1}{c}{0.099}      &  0.721      \\ 
			DMM     & LeNet    & 0.18     & \multicolumn{1}{c}{0.620}      &  0.755      & \multicolumn{1}{c}{0.619}      &  0.758      \\ 
			Anole     & LeNet   & 0.18     & \multicolumn{1}{c}{\textbf{0.822}} &  \textbf{0.844} & \multicolumn{1}{c}{\textbf{0.821}} &  \textbf{0.846} \\ \bottomrule
		\end{tabular}
	}
\end{table}

\begin{table}[]
	\caption{Performance of different methods on the CGD task.}
	\label{table:cgd}
	\centering
    \begin{threeparttable}[]
        \begin{tabular}{c|ccccc}
                \hline
                Method$^1$ & Latency & ER   & MAE  & MSE   & PCCS   \\ \hline
                SSM & 0.87s & 2.28 & 1.48 & 2.69  & 0.17 \\
                CDG & 0.87s & 5.05 & 3.27 & 14.16 & 0.12 \\
                DMM & 0.90s & 2.14 & 1.38 & 2.34  & 0.16 \\
                Anole & 1.05s  & \textbf{1.38} & \textbf{0.90} & \textbf{1.40}  & \textbf{0.40} \\ \hline
        \end{tabular}
        \begin{tablenotes}
            \item[1] There is no effective deep model (\emph{i.e.}, SDM) so far for the CGD task owing to the lacking of large-scale datasets.
        \end{tablenotes}
    \end{threeparttable}
\end{table} 

\subsection{Cross-scene Experiments}
\label{sec:accuracy}

In this experiment, we investigate the performance of all candidate methods cross fast-changing scenes, using samples in the test set of seen data. For the VD task, F1 score is calculated every ten frames to show the instantaneous performance changes.

\subsubsection{Performance Comparison} 
\begin{figure*}[]
	\centering
	\subfigure[KITTI]{
		\begin{minipage}[b]{0.27\textwidth}
			\centering
			\includegraphics[height=3.5cm]{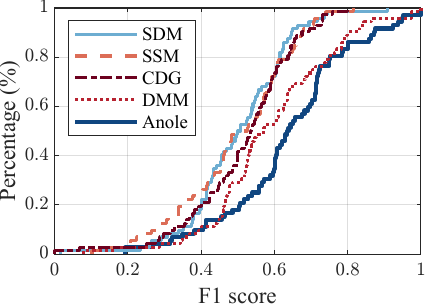}
		\end{minipage}
		\label{fig:f1_kitti}
	}
	\hspace{0.4cm}
	\subfigure[BDD100k]{
		\begin{minipage}[b]{0.27\textwidth}
			\centering
			\includegraphics[height=3.5cm]{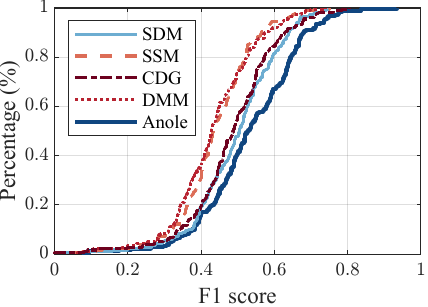}
		\end{minipage}
		\label{fig:f1_bdd}
	}
	\hspace{0.4cm}
	\subfigure[SHD]{
		\begin{minipage}[b]{0.27\textwidth}
			\centering
			\includegraphics[height=3.5cm]{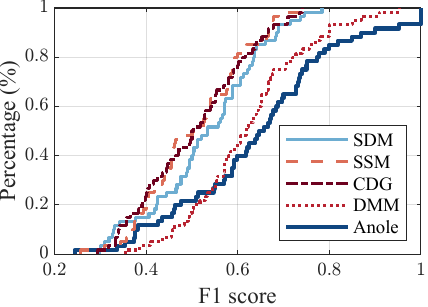}
		\end{minipage}
		\label{fig:f1_real}
	}
	\caption{CDFs of F1 score of all candidate methods on each source dataset, demonstrating the advantage of Anole over candidate methods, including the versatile large SDM. Note that the more the line leans towards the \textbf{bottom right corner}, the better the performance.}
	\label{fig:cdf_overall}
\end{figure*}

Table \ref{tab:mnist} lists the accuracy of all candidate methods over all cross-scene HDR test images.
Table \ref{table:cgd} list the considered metrics of all candidate methods on the testing samples of CGD task.
Figure \ref{fig:cdf_overall} plots the CDFs of F1 score of all candidate methods on each test set of seen data selected from KITTI, BDD100k and SHD, respectively.
For both tasks, Anole outwits other methods in terms of accuracy. For example, on the HDR task, Anole can even outwits SDM by 6.3\%; on the CGD task, Anole can achieve a prediction PCCS increase by 135.5\%; on the VD task, ANole can outperform SDM in terms of SDM by over 15\%. Moreover, other methods exhibit inconsistent performance across different datasets. For example of the VD task, DMM gains good performance on the KITTI and SHD datasets, while SDM only performs well on the BDD100K dataset. This discrepancy arises because DMM fits simpler datasets whereas SDM is biased towards BDD100k due to the overwhelming number of training samples.

\subsubsection{Effect of Data Segmentation and Model Adaptation}
It is a common practice to train an individual model on each datasets (\emph{i.e.}, DMM) or to segment a dataset according to feature similarity and train respective models (\emph{i.e.}, CDG). It can be seen that DMM performs similarly to Anole for simple training datasets such as all MNIST datasets for the HDR task, and the KITTI and the SHD datasets for the VD task, but DMM performs poorly on large and complex datasets like BDD100k. In contrast, CDG trains and selects models on similar data samples. However, the inference accuracy of CDG is not as good as that of Anole over all test sets for both tasks. This demonstrates Proposition~\ref{proposition1}, which states that a model trained on a scene may not always perform well on that scene. In contrast, Anole employs a decision model to learn the appropriate scenes and determines which model is most suitable for online prediction, resulting in stable performance.
Furthermore, although deep-model-based method SDM is generally assumed to have better performance, we surprisingly find that Anole outwits SDM on all test sets. This implies that training a single large DNN model for cross-scene inference is more difficult than training and choosing from a set of specialized compressed models.

\begin{table*}[]
	\caption{Inference accuracy of all candidate methods obtained on unseen data. The best results are indicated in bold while the second-best results are marked in blue.}
	\label{tab:unseen}
	\centering
	\scalebox{1}{
		\begin{threeparttable}[]
			
			\begin{tabular}{c|cccccc|c}
				\toprule
				\multirow{2}{*}{Method} & \multicolumn{1}{c|} {KITTI} &  \multicolumn{4}{c|} {BDD100k} &  \multicolumn{1}{c|} {SHD} & \multirow{2}{*}{Average} \\ \cline{2-7}
				&  \multicolumn{1}{c|}{Street, Day}  & Urban, Night & Urban, Day & Highway, Dusk  & Street, Night  &  \multicolumn{1}{|c|}{Tunnel, Night}  &                       \\ \hline \hline
				SDM$^1$   &  \multicolumn{1}{c|}{0.437} & 0.531 & \textbf{0.477} & \textbf{0.476} & \textbf{0.468} &  \multicolumn{1}{|c|}{\color{blue}{0.409}} & \color{blue}{0.466} \\ \hline
				SSM   &  \multicolumn{1}{c|}{0.387} & 0.514 & 0.335 & 0.404 & 0.454 &  \multicolumn{1}{|c|}{0.370} & 0.411 \\ 
				CDG   & \multicolumn{1}{c|}{\color{blue}{0.459}} & \color{blue}{0.537} & 0.453 & 0.410 & 0.440 &  \multicolumn{1}{|c|}{0.401} & 0.450 \\ 
				DMM    & \multicolumn{1}{c|}{0.407} & 0.482 & 0.382 & 0.388 & 0.384 &  \multicolumn{1}{|c|}{0.374} & 0.403 \\ 
				Anole &  \multicolumn{1}{c|}{\textbf{0.506}} & \textbf{0.590} & \color{blue}{0.453} & \color{blue}{0.440} & \color{blue}{0.461} &  \multicolumn{1}{|c|}{\textbf{0.470}} & \textbf{0.487} \\ \bottomrule
			\end{tabular}
			\begin{tablenotes}
				\item[1] SDM uses a deep model, resulting in higher latency, larger memory usage (Table \ref{tab:latency}), and higher power consumption (Figure \ref{fig:sys}).
			\end{tablenotes}
		\end{threeparttable}
	}
\end{table*}

\subsection{New-scene Experiments}
In this experiment, we examine the performance of all candidate methods in new scenes, using unseen data. Particularly, for the VD task, six unseen video clips include one clip from KITTI with attributes of \{\emph{Street}, \emph{Day}\}, four scenes from BDD100k with attributes of \{\emph{Urban}, \emph{Night}\}, \{\emph{Urban}, \emph{Day}\}, \{\emph{Highway}, \emph{Dusk}\}, and \{\emph{Street}, \emph{Night}\}, and one scene from SHD with attributes of \{\emph{Tunnel}, \emph{Night}\}. 
Table \ref{tab:mnist} and Table~\ref{tab:unseen} list the accuracy results for the HDR and VD tasks, respectively.
It can be seen that though SDM with a much larger model size is expected to excel other shallow-model-based methods on unseen scenes, Anole demonstrates supreme generalization ability and even outperforms SDM on all unseen data. As for unseen scenes from BDD100k, Anole can still achieve high accuracy comparable to that of SDM.

\subsection{Real-world Experiments}

\begin{figure}[]
	\centering
	\subfigure[Implementation of Anole on Jetson TX2 NX deployed on a UAV]{
		\begin{minipage}[t]{0.22\textwidth}
			\centering
			\includegraphics[height=2.95cm]{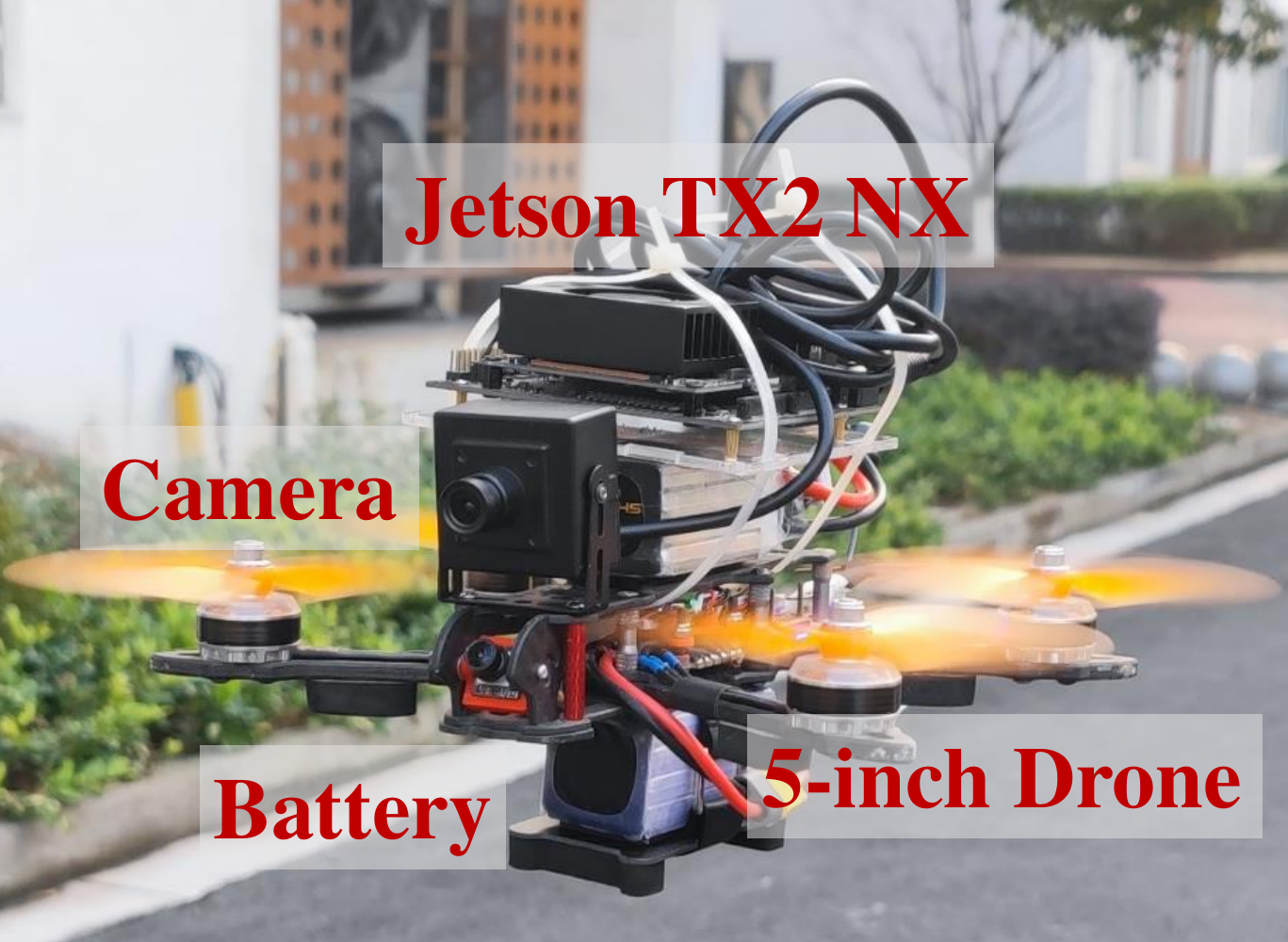}
		\end{minipage}
		\label{fig:implementation}
	}
	\subfigure[Results in a typical night scenario]{
		\begin{minipage}[t]{0.22\textwidth}
			\centering
			\includegraphics[height=2.95cm]{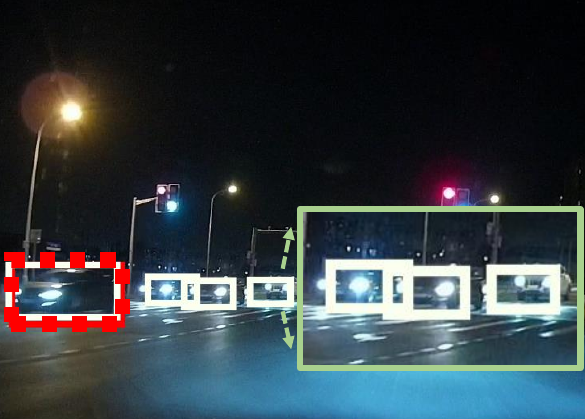}
		\end{minipage}
		\label{fig:visualResult}
	}
	\caption{(a) Implementation of Anole on a Jetson TX2 NX connected with a 1080p HD camera. (b) Visualization of vehicles detected in a night scenario.}
\end{figure}

\begin{figure}[]
	\centering
	\includegraphics[height=4.3cm]{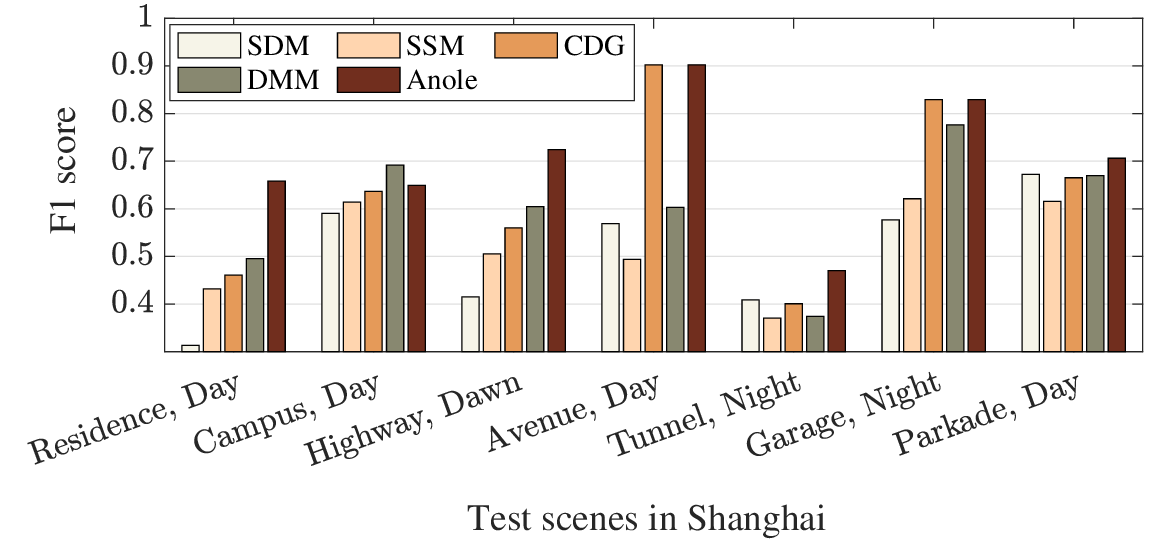}
	\caption{F1 score of all methods on test scenes in Shanghai, where Anole exceeds other methods with a latency of less than 20 ms on Jetson TX2 NX.}
	\label{fig:bar_real}
\end{figure}

As depicted in Figure \ref{fig:implementation}, we implement all methods on the Nvidia Jetson TX2 NX connected with a 1080p HD camera to conduct real-world experiments in Shanghai city. Well-trained compressed models and the decision model are downloaded to the Jetson device.  We conduct real-world experiments in seven driving scenarios with different road conditions and different time in a day.  LabelImg \cite{tzutalin2015labelimg} is used to label all recorded frames as the ground truth for offline analysis.
Figure \ref{fig:bar_real} plots the F1 score of all methods. Anole outperforms all other candidate methods in all test scenarios. 
We visualize the car detection results of Anole (white solid frames) and SDM (red dashed frames) in a typical night driving scenario in Figure~\ref{fig:visualResult}. The inference results obtained using SDM frequently contain errors, especially false negative errors as shown in the enlarged subgraph.

\begin{table}[]
	\caption{Inference latency and memory consumption on mobile devices.}
	\label{tab:latency}
	\centering
	\scalebox{1}{
		\begin{threeparttable}
			\begin{tabular}{c|ccc|cc}
				\toprule
				\multirow{2}{*}{\diagbox{Model}{Metric}}  & \multicolumn{3}{c|}{Latency (ms)}                                                        & \multicolumn{2}{c}{GPU Memory (MB)}                              \\ \cline{2-6} 
				& \multicolumn{1}{c|}{Nano}    & \multicolumn{1}{c|}{TX2} & \multicolumn{1}{c|}{Lap.} & \multicolumn{1}{c|}{Loading} & \multicolumn{1}{c}{Execution} \\ \hline \hline
				$\mathcal{M}_{sce.}$ + $\mathcal{M}_{dec.}$ & \multicolumn{1}{c|}{23.2}     & \multicolumn{1}{c|}{3.1}    & 20.8                         & \multicolumn{1}{c|}{44}   & 584                          \\ \hline
				YOLOv3                   & \multicolumn{1}{c|}{313.8} & \multicolumn{1}{c|}{42.9}  & 62.2                        & \multicolumn{1}{c|}{240$\times n$\tnote{1}}   & 1,730                         \\ \hline
				YOLOv3-tiny              & \multicolumn{1}{c|}{37.8}  & \multicolumn{1}{c|}{10.8}   & 32.2                        & \multicolumn{1}{c|}{40$\times n$}   & 1,120                         \\ \bottomrule
			\end{tabular}
			\begin{tablenotes}
				\item[1] $n$ denotes the number of compressed models to load.
			\end{tablenotes}
	\end{threeparttable}}
\end{table}

\subsection{Inference Latency} We evaluate the inference latency of the decision model $\mathcal{M}_{decision}$ and compressed models on different mobile devices. The results are shown in Table~\ref{tab:latency}. The results reveal that YOLOv3-tiny exhibits significantly lower latency when compared to deep YOLOv3, which is generally deemed unsuitable for deployment on devices. For instance, the latency of YOLOv3-tiny on Jetson Nano is 87.9\% lower than that of YOLOv3. This highlights the substantial potential for accelerating inference using shallow models.
It is also evident that $\mathcal{M}_{decision}$ can be executed in real-time on embedded mobile devices such as Jetson Nano, with a latency as low as 23.2 ms, making it suitable for online inference applications.



\subsection{Memory and Power Consumptions}
\begin{figure}[]
	\centering
	\includegraphics[height=2.7cm]{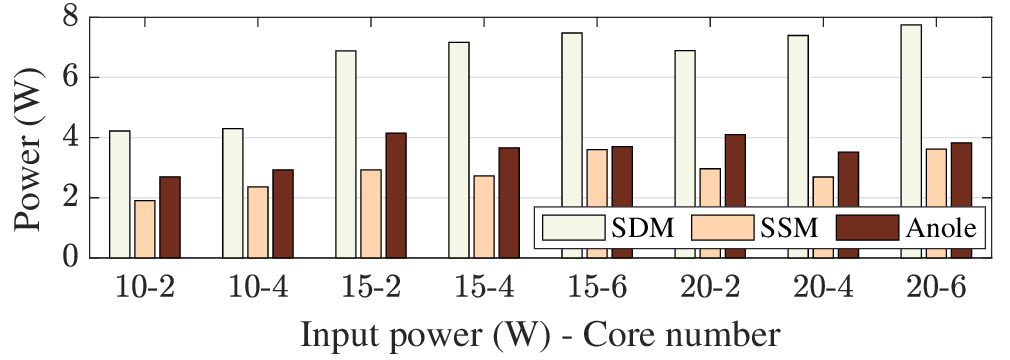}\\
	\includegraphics[height=2.7cm]{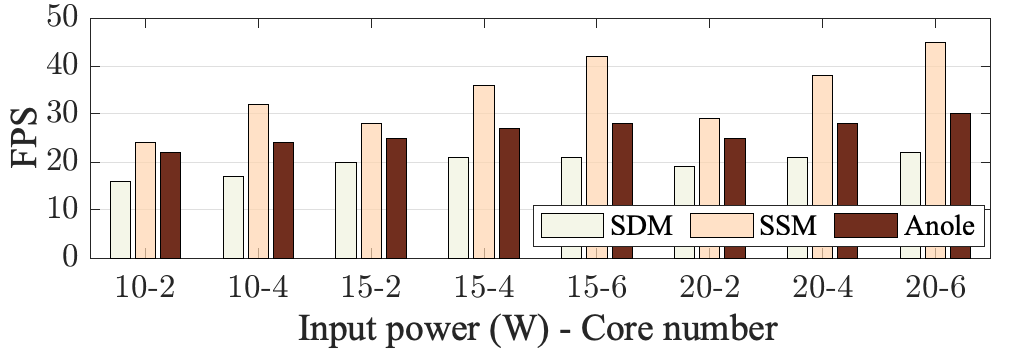}
	\caption{Power consumption and inference speed of different methods in various power modes.}
	\label{fig:sys}
\end{figure}

We investigate the memory consumption of different models from the following two aspects, \emph{i.e.}, loading model only, and the memory consumption during inference with a batch size of 1. Table~\ref{tab:latency} demonstrates that memory consumption for loading model is significantly lower than that during inference, owing to the presence of hidden parameters during inference. 
We also examine the impact of different power configurations adopted by Jetson TX2 NX to the performance of Anole. The power consumption and inference speed of Anole and baselines under different power modes are shown in Figure \ref{fig:sys}, respectively. Anole achieves a 45.1\% reduction in power consumption compared with SDM and a inference speed of over 30 FPS with an input power of 20W running 6 cores.

\label{sec:system}

\section{Related Work}

\subsection{DNN Prediction on Mobile Devices}
To perform DNN inference on mobile devices, new DNNs are specially designed~\cite{howard2017mobilenets, zhang2018shufflenet, sanh2019distilbert, jiao2019tinybert} or existing DNNs are compressed to match the computing capability of a mobile device~\cite{han2015learning, wang2020pruning}. First, model structure can be optimized to reduce complexity~\cite{chen2020addernet, howard2017mobilenets, zhang2018shufflenet}. Second, quantization precision can be reduced to minimize computational cost, \emph{e.g.}, use integers instead of floating-point numbers~\cite{alibaba2020mnn, elsen2020fast}. Third, the neural network model can also be accelerated by pruning, \emph{i.e.}, deleting some neurons in the neural network~\cite{han2015learning, wang2020pruning, liu2018rethinking}. Scene information is also utilized for model compression on edge/mobile devices \cite{feng2021palleon, xu2022litereconfig, wang2022enabling, jiang2021flexible}. Finally, model distillation can distill the knowledge of large models into small models~\cite{sanh2019distilbert, jiao2019tinybert}. Such schemes ensure real-time model inference at the expense of compromised accuracy.

Another direction is to divide DNNs and perform collaborative inference on both edge devices and the cloud~\cite{hu2019dynamic, kang2017neurosurgeon, banitalebi2021auto, zheng2022towards, laskaridis2020spinn, guo2021mistify}, or to transmit compressed sensory data to the cloud for data recovery and model inference~\cite{Liu2019, zhang2021elf}. 
Neurosurgeon~\cite{kang2017neurosurgeon} partitions the computation of each DNN inference task in a layer granularity. CLIO \cite{huang2020clio} addresses the instability of network conditions and optimizes inference under different network states. 
These approaches need coordination with the cloud for each inference, leading to unpredictable inference delays when the communication link is unstable or disconnected.
However, they prove inadequate for cross-scene mobile inference scenarios where even deep models are unable to cope.

\subsection{Cross-scene DNN Prediction}

Data-driven machine learning models face challenges in maintaining robust inference performance when dealing with cross-scene inference \cite{zhai2021rise}. One natural approach for scene partitioning is to partition the scene based on prior knowledge or historical samples. First, based on prior knowledge, a similarity graph is constructed to cluster similar domains together \cite{han2014encoding}. However, obtaining such prior knowledge based on domain expertise can be challenging. Second, the original data or their extracted features can be utilized for more automated scene partitioning \cite{lu2022out}. However, these methods may result in the loss of critical information in complex systems \cite{zheng2019metadata}. Cross-scene DNN prediction can also be enhanced given a golden model in the cloud for online sample labelling \cite{khani2023recl, bhardwaj2022ekya}. However, the existence of a perfect or golden model is not always feasible. 

\subsection{Mixture of Experts}
In recent years, we have witnessed the success of Mixture of Experts (MoE) \cite{jacobs1991adaptive, shazeer2016outrageously}, especially in efficient training of large language models (LLM). MoE employs multiple experts for model training, each for one domain. Then, a gate network will be used to determine the correspondence between samples and experts. Though inspired by MoE, Anole differs from MoE in the following 2 aspects. First, experts in MoE are diversified by constraints of losses, but they themselves cannot be related to the scene. In fact, the main purpose of MoE is to expand the number of model parameters, rather than to customize and select scene-specific models. Second, MoE is just a model architecture, and models based on MoE architecture still need to deploy the entire model during deployment. Therefore, MoE-based models often require a significant amount of memory, which is unacceptable for mobile agents like UAVs. In contrast, Anole employs multiple compressed models for online model inference, each designed for one scene. Only a few compressed models are needed to be deployed during online inference. Therefore, Anole is more suitable for mobile devices only with limited resources.

\section{Discussion}

\subsection{How to choose the CMT-DM and CMT-SM algorithm?}

The effect of Anole relies on an army of compact DNN models, which can be given in advance or trained by the two algorithms (\emph{i.e.}, Alg.\ref{Alg:CMT_DM}, CMT-DM and Alg.\ref{Alg:CMT_SM}, CMT-SM) proposed in Section \ref{sec:cmt}. We compare the differences of CMT-DM and CMT-SM, which can guide the choice of the two models training algorithms. 

First, in general, CMT-SM is better in terms of inference accuracy than CMT-DM because CMT-SM combines multiple scenes based on both semantics and model training performances, while CMT-SM only considers the similarity of semantics of different scenes. Therefore, with adequate computing resources, we should first choose the CMT-SM algorithm. 

However, each combination of scenes requires one round of model training. In fact, the upper bound complexity of the CMT-SM algorithm can be computed by considering the case where no combination of tasks is conducted and the number of scenes never reduces. In this case, the complexity of CMT-SM with $n$ scenes at the $i$th iteration is $O(T_R \cdot (n-i+1))$, where $T_R$ denotes the time for conducting a training session. Thus, the overall complexity $\mathcal{O}(T_R \cdot  (\sum_1^{n-1} i)) \in \mathcal{O}(T_R \cdot (n \cdot (n-1) /2)) \in \mathcal{O}(T_R \cdot n^2)$. Such a complexity is too large if $T_R$ is a large number, \emph{e.g.}, the training time of a typical DNN. In these cases (\emph{e.g.}, the HDR task and the VD task), choosing the CMT-DM algorithm is a practical option.

\subsection{Limitation of Anole}

Anole has the limitation that, Anole may encounter new scenes which it cannot handle. The performance of Anole relies on a dataset containing all possible scenes and the goal of Anole is to achieve stable model inference across the seen scenes and try our best to generalize to new scenes. The reason why Anole may generalize to unseen new scenes is that the unseen new scenes may be similar to the seen scenes in the view of compressed models. However, we cannot always guarantee this. There may be some new scenes that Anole cannot handle. Fortunately, experiments have shown that such situations are not common. When encountering new scenes that cannot be processed, Anole can make timely reminders and switch to the large model on the cloud to ensure basic inference response.

\subsection{Can Mobile Foundation Model (FM) replace Anole?}

Mobile Foundation Model \cite{meng2024} inspired by Large Language Model (LLM) such as ChatGPT also proves to be potential to solve the cross-scene online model inference (OMI) problem on mobile devices. Mobile FM also shows the amazing generalization performance on typical mobile tasks. However, Anole can exceed the mobile FM in the following 2 aspects. First, Anole has extremely low memory requirements. Anole only needs to load a cache of compressed models to memory or GPU memory (usually $<$ 2GB), while mobile FM needs to load the whole FM to mobile devices ($>$ 7.5 GB). As a result, Anole is more suitable for the low-end devices with limited memories. Second, the power consumption of Anole is much less than mobile FM. Despite multiple models are utilized in Anole, each time Anole only use one model for inference. On the contrary, mobile FM needs a whole inference of a large neural network, consuming tens of times more power than Anole. Therefore, Anole is also more suitable for the battery-powered mobile devices such as UAVs. Second, mobile FM cannot work well on all tasks on mobile devices. Without scene-related information, for tasks with less well-annotated data (such as the CGD task), mobile FM cannot solve these task well.

\section{Conclusion}

In this paper, we have proposed Anole, an online model inference scheme on mobile devices. Anole employs a rich set of compressed models trained on a wide variety of human-defined scenes and offline learns the implicit mode-defined scenes characterized by these compressed models via a decision model. Moreover, the most suitable compressed models can be dynamically identified according to the current testing samples and used for online model inference. As a result, Anole can deal with unseen samples, mitigating the impact of OOD problem to the reliable inference of statistical models. Anole is lightweight and does not need network connection during online inference. It can be easily implemented on various mobile devices at a low cost. Extensive experiment results demonstrate that Anole can achieve the best inference accuracy at a low latency.

\bibliographystyle{IEEEtran}
\bibliography{ref}

\begin{IEEEbiography}[{\includegraphics[width=1in,height=1.25in,keepaspectratio]{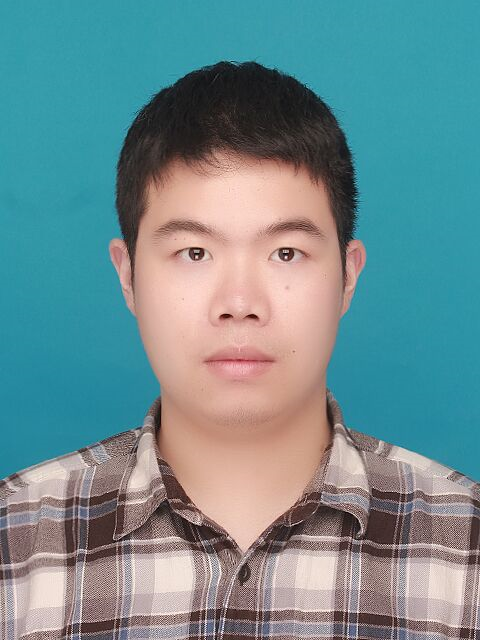}}]{Yunzhe Li} received his B.S. degree in computer science from Wuhan University in 2021. He is a PhD student in the Department of Computer Science and Engineering, Shanghai Jiao Tong University. His research interests include mobile sensing and edge computing. He has published papers in ICDCS, CIKM, NeurIPS, TMC, TC, etc. He received the Rising Star Award in 2023 from the KubeEdge of Cloud Native Computing Foundation.
\end{IEEEbiography}

\vspace{0.5cm}

\begin{IEEEbiography}[{\includegraphics[width=1in,height=1.25in,keepaspectratio]{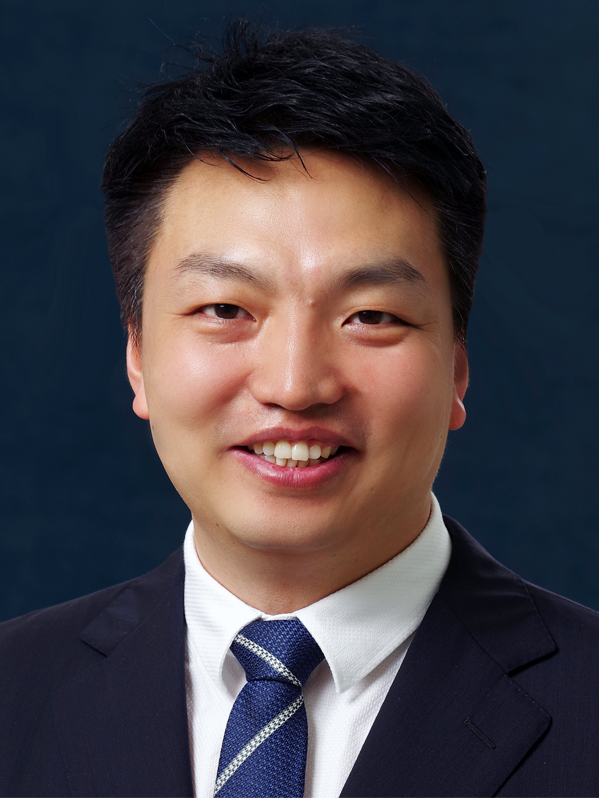}}]{Hongzi Zhu}
	received the PhD degree in computer science from Shanghai Jiao Tong University, in 2009. He was a post-doctoral fellow with the Department of Computer Science and Engineering, Hong Kong University of Science and Technology, and the Department of Electrical and Computer Engineering, University of Waterloo, in 2009 and 2010, respectively. He is now a professor with the Department of Computer Science and Engineering, Shanghai Jiao Tong University. His research interests include mobile sensing, mobile computing, and Internet of Things. He received the Best Paper Award from IEEE Globecom 2016. He is an associate editor for the IEEE Transactions on Vehicular Technology and the IEEE Internet of Things Journal. He is an IEEE Senior Member. For more information, please visit http://lion.sjtu.edu.cn.
\end{IEEEbiography}

\vspace{0.5cm}

\begin{IEEEbiography}[{\includegraphics[width=1in,height=1.25in,keepaspectratio]{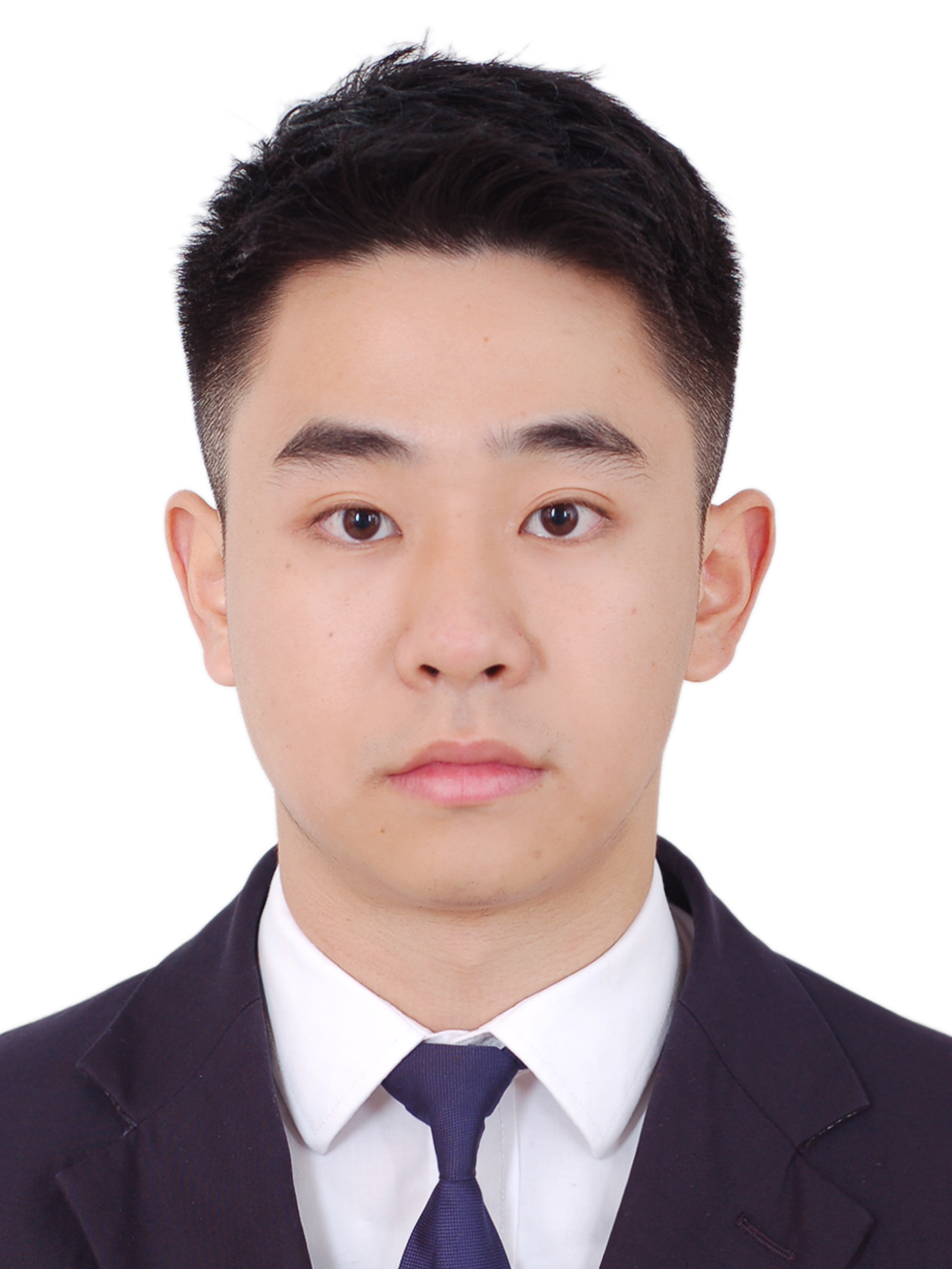}}]{Zhuohong Deng} received the B.E. degree in Computer Science and Technology from Beijing Forestry University in China, in 2021. He is currently working toward the Master degree in the Department of Computer Science and Engineering at Shanghai Jiao Tong University. His research interest is artificial intelligence system.
\end{IEEEbiography}

\vspace{0.5cm}

\begin{IEEEbiography}[{\includegraphics[width=1in,height=1.25in,keepaspectratio]{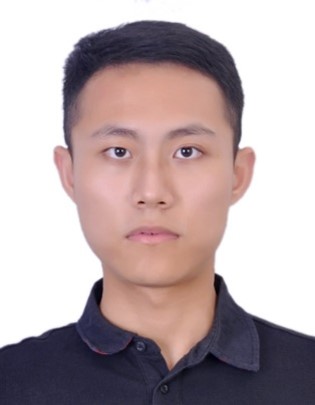}}]
	{Yunlong Cheng} received the B.E. degree in Computer Science and Technology from Shanghai Jiao Tong University, China, in 2021. Presently, he is pursuing a PhD in Computer Science and Technology at Shanghai Jiao Tong University. His research focuses on data mining, prediction, and scheduling algorithms within the context of cloud services. He has published papers in IPDPS, TCS, DASFAA, etc.
\end{IEEEbiography}

\vspace{0.5cm}

\begin{IEEEbiography}[{\includegraphics[width=1in,height=1.25in,keepaspectratio]{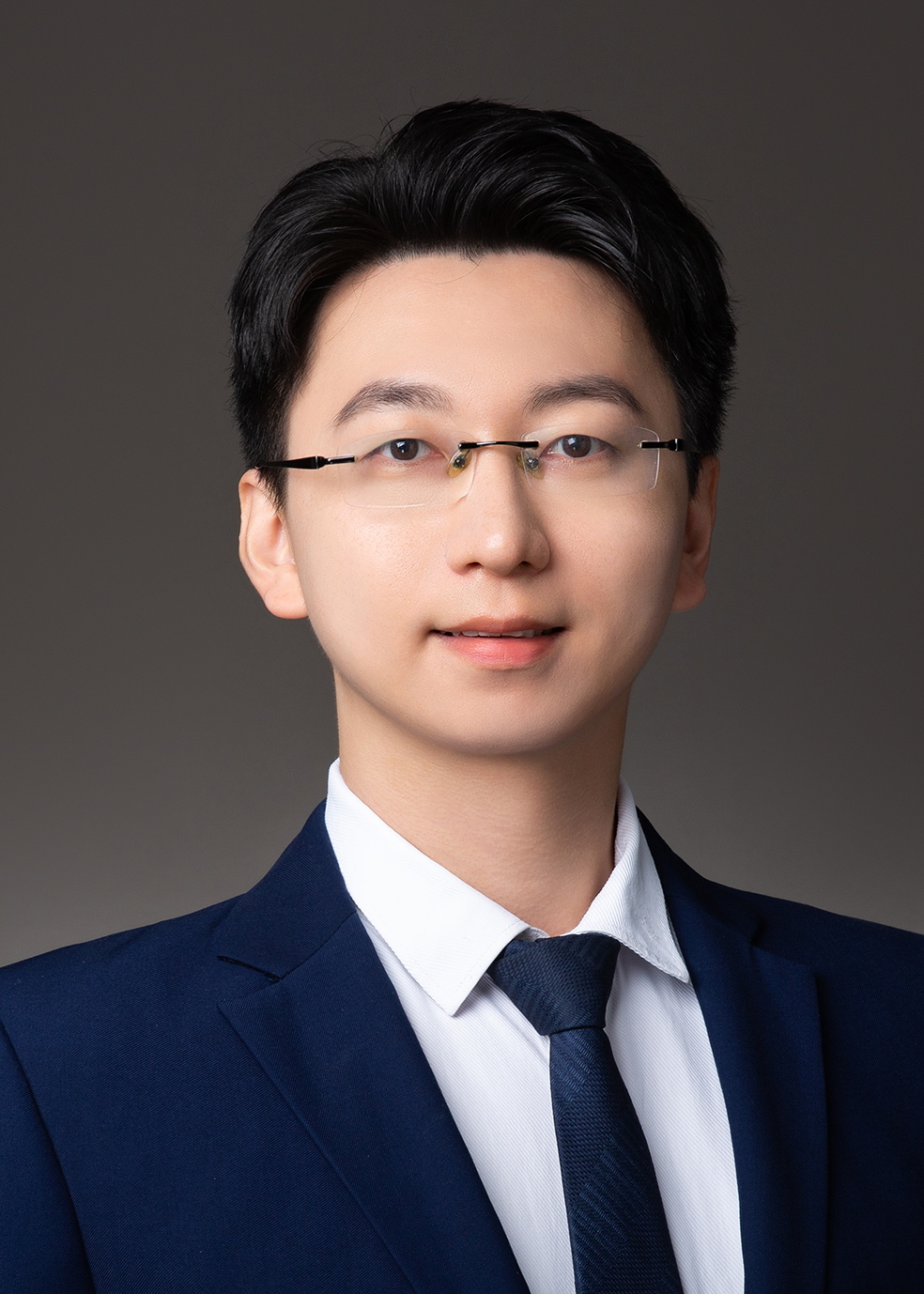}}]
{Zimu Zheng} is currently a head research engineer at Huawei Cloud. He received his B.Eng. degree from South China University of Technology and his Ph.D degree from the Hong Kong Polytechnic University. His research interest lies in edge intelligence, benchmarking, multi-task learning, and AIoT. He received the Best Paper Award of ACM e-Energy 2018 and the Best Paper Award of ACM BuildSys 2018. He is an IEEE member with publications in IEEE TPDS, IEEE TPAMI, IEEE IOTJ, IEEE INFOCOM, IEEE ICDCS, IEEE TSUSC, etc. He has also received several awards for outstanding technical contributions in Huawei and serves as a co-chair in the KubeEdge SIG AI of Cloud Native Computing Foundation. 
\end{IEEEbiography}

\vspace{0.5cm}

\begin{IEEEbiography}[{\includegraphics[width=1in,height=1.25in,keepaspectratio]{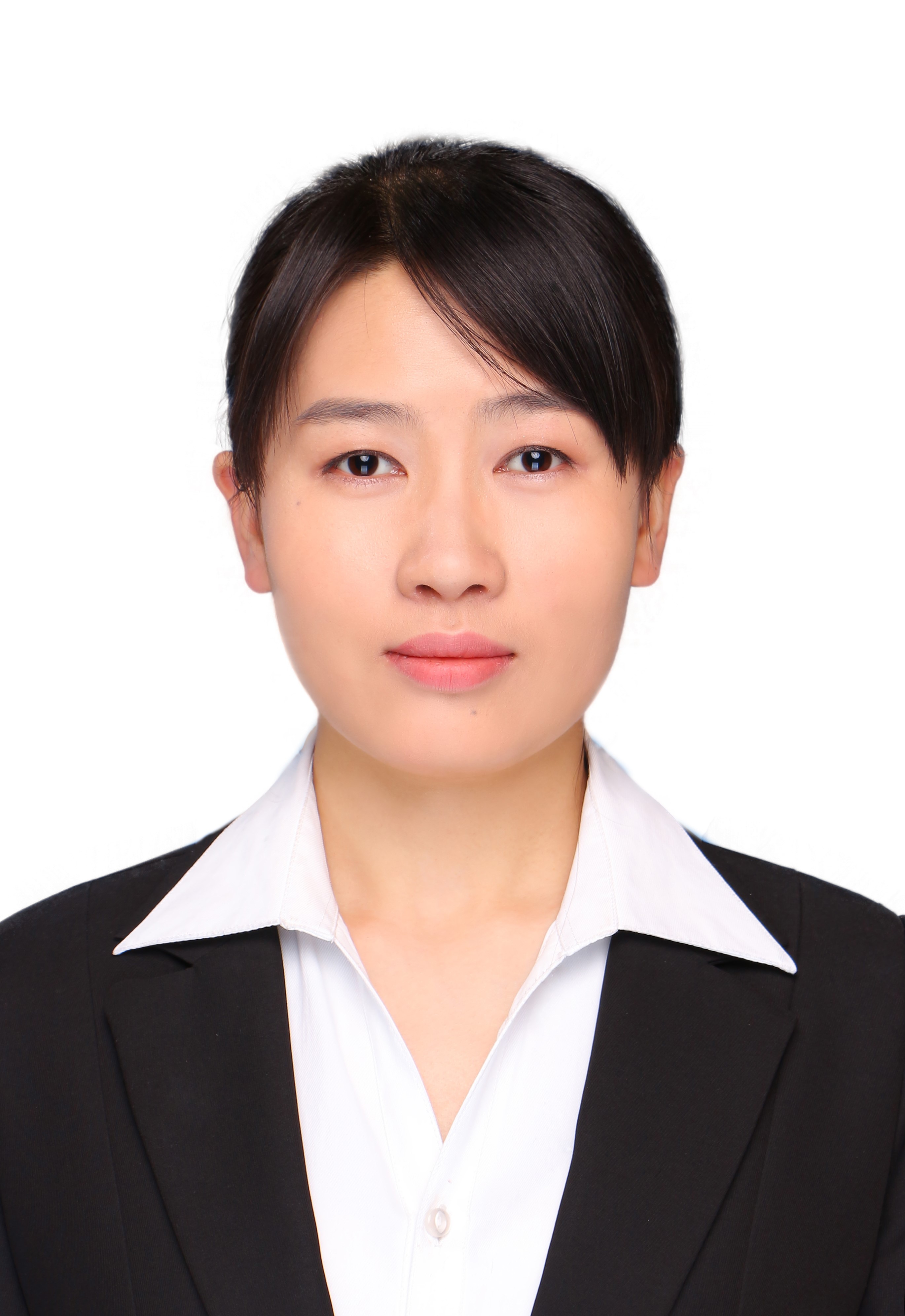}}]
	{Liang Zhang}
	received the B.E. degree in Computer Science and Technology from Northeastern University in China, in 2018. She is currently working toward the PhD degree in the Department of Computer Science and Engineering at Shanghai Jiao Tong University. Her research interest includes stream processing and resource scheduling in the cloud or edge computing environment. For more information, please visit https://zl-cs.github.io/.
\end{IEEEbiography}

\vspace{0.5cm}

\begin{IEEEbiography}[{\includegraphics[width=1in,height=1.25in,keepaspectratio]{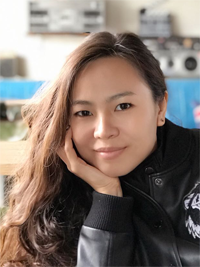}}]
	{Shan Chang}
	received the PhD degree in computersoftware and theory from Xian Jiaotong University,in 2013. From 2009 to 2010, she was a visitingscholar with the Department of Computer Scienceand Engineering, Hong Kong University of Scienceand Technology. She was also a visiting scholar withthe BBCR Research Lab, University of Waterloo,from 2010 to 2011. She is now an associate professor with the Department of Computer Scienceand Technology, Donghua University, Shanghai. Herresearch interests include security and privacy in mobile networks and sensor networks. She is a member of the IEEE.
\end{IEEEbiography}

\vspace{0.5cm}

\begin{IEEEbiography}[{\includegraphics[width=1in,height=1.25in,keepaspectratio]{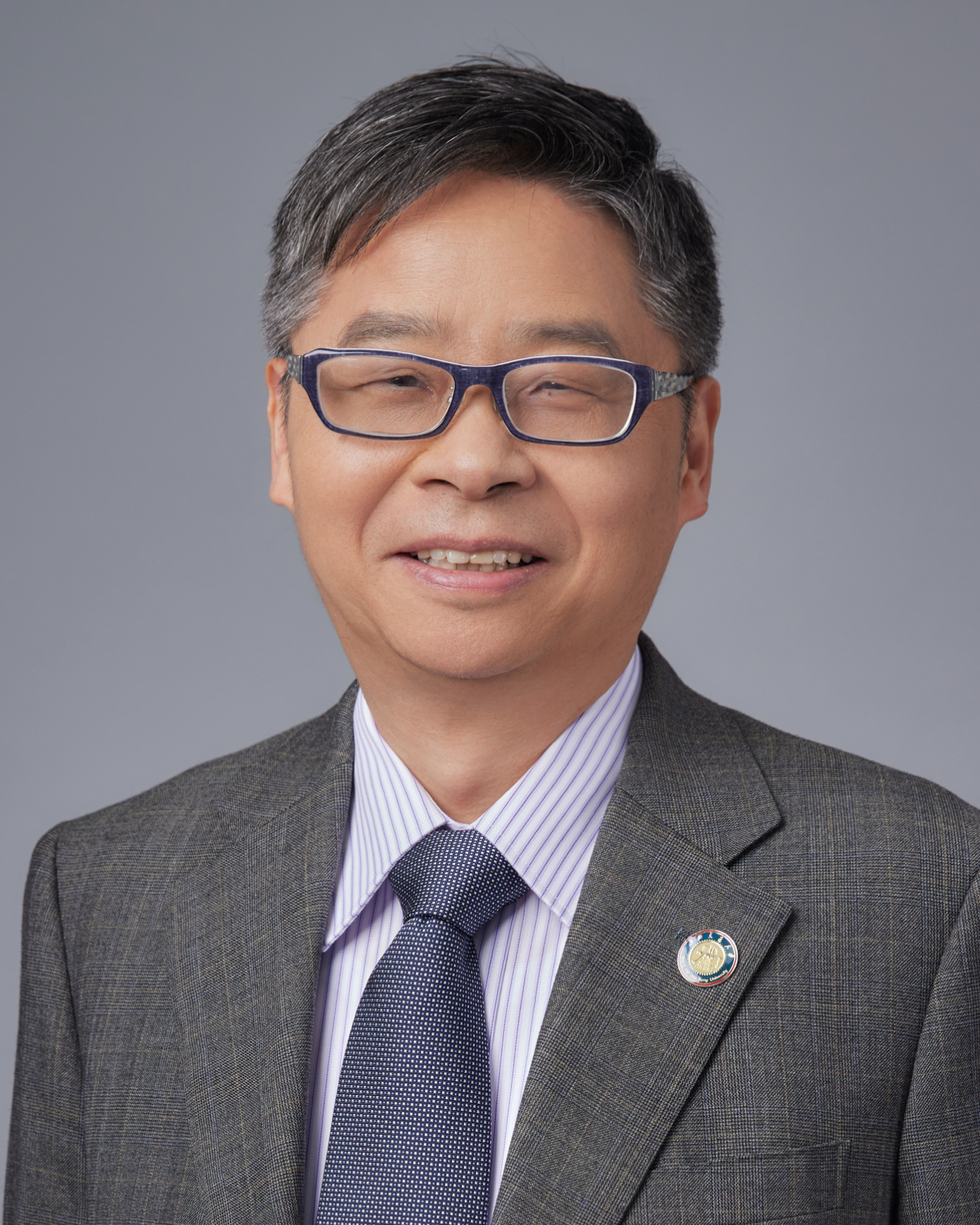}}]{Minyi Guo} is a Zhiyuan Chair Professor in the Department of Computer Science and Engineering at Shanghai Jiao Tong University, China. He received the PhD degree
in computer science from the University of Tsukuba, Japan. He is currently Zhiyuan Chair professor. His present research interests include parallel/distributed computing, compiler optimizations, embedded systems, pervasive computing, big data and cloud computing. He is now on the editorial board for IEEE Transactions on Parallel and Distributed Systems, IEEE Transactions on Cloud Computing and Journal of Parallel and Distributed Computing. He is a fellow of IEEE. 
\end{IEEEbiography}

\balance

\vfill

\end{document}